\documentclass[final,3p,number,sort&compress]{elsarticle}

\usepackage{geometry} 
\usepackage{amsmath,amssymb,amsthm} 
\usepackage[american]{babel}
\usepackage{microtype} 
\usepackage{algorithm} 
\usepackage{algorithmic}

\newtheorem{theorem}{Theorem}

\usepackage{enumitem} 
\usepackage{url} 

\usepackage{booktabs} 
\usepackage{multirow} 
\usepackage{array} 
\usepackage{makecell} 
\usepackage{multirow} 
\usepackage{makecell} 

\usepackage{graphicx} 
\usepackage{subfig}

\usepackage{caption}
\makeatletter 
\def\@makecaption#1#2{%
	\vskip\abovecaptionskip
	\sbox\@tempboxa{#1 #2}%
	{\bfseries #1} #2\par
	\vskip\belowcaptionskip}
\makeatother
\usepackage[justification=centering]{caption}
\usepackage[capitalize]{cleveref} 

\newcommand{\figref}[1]{Fig.~\ref{#1}} 
\usepackage{cases} 
\usepackage{float} 

\journal{}

\begin{document}
\captionsetup[figure]{labelfont={bf},labelsep=period,name={Fig.}} 
	\begin{frontmatter}
		
		
		
		\title{Multi-channel Nuclear Norm Minus Frobenius Norm Minimization for Color Image Denoising}
		

		\author{Yiwen Shan}
		\author{Dong Hu}
		\author{Zhi Wang\corref{cor}}
		\ead{chiw@swu.edu.cn}
		\cortext[cor]{Corresponding author}
		
		\author{Tao Jia}
		\address{
			College of Computer and Information Science,
			Southwest University,
			Chongqing,
			400715,
			PR China
		}
		\begin{abstract}
		Color image denoising is frequently encountered in various image processing and computer vision tasks. 
		One traditional strategy is to convert the RGB image to a less correlated color space and denoise each channel of the new space separately. However, such a strategy can not fully exploit the correlated information between channels and is inadequate to obtain satisfactory results.
		To address this issue, this paper proposes a new multi-channel optimization model for color image denoising under the nuclear norm minus Frobenius norm minimization framework. 
		Specifically, based on the block-matching, the color image is decomposed into overlapping RGB patches. For each patch, we stack its similar neighbors to form the corresponding patch matrix. The proposed model is performed on the patch matrix to recover its noise-free version. During the recovery process, a) a weight matrix is introduced to fully utilize the noise difference between channels; b) the singular values are shrunk adaptively without additionally assigning weights. With them, the proposed model can achieve promising results while keeping simplicity.
        To solve the proposed model, an accurate and effective algorithm is built based on the alternating direction method of multipliers framework. The solution of each updating step can be analytically expressed in closed-from. 
        Rigorous theoretical analysis proves the solution sequences generated by the proposed algorithm converge to their respective stationary points. 
        Experimental results on both synthetic and real noise datasets demonstrate the proposed model outperforms state-of-the-art models.
		\end{abstract}
		
		
		
		
	\end{frontmatter}
	
	\section{Introduction}
	The image denoising problem, which is to recover the underlying clean image from its noisy observation, has generated considerable research interest in recent years. Such a task is a challenging but fundamental problem and can be found in many image processing and computer vision tasks, such as image segmentation \cite{iSegment1, iSegment2}, remote sensing imaging \cite{RemoteSensingImaging1, RemoteSensingImaging2}, object recognition \cite{objRecognition}, and video denoising \cite{videoDenoising}. Mathmatically, image denoising problem can be formulated as 
	\begin{equation}\mathbf{y} = \mathbf{x} + \mathbf{n},
	\end{equation}
	where $\mathbf{y}$, $\mathbf{x}$ and $\mathbf{n}$ are the noisy observation, the clean image and the addictive white Gaussian noise (AWGN), respectively. During the past decade, lots of methods have been proposed to solve this problem, which can be roughly catagorized as transform domain methods \cite{BM3D, CBM3D}, low-rank minimization methods \cite{WNNM, WSNM, MCWNNM, MCWSNM} and CNN-based methods \cite{DnCNN, FFDNet}. 
	\par 
	Among them, low-rank minimization methods have boosted the denoising performance significantly. The goal of low-rank minimization \cite{wang2,wang3,wang4} is to find a matrix with minimum rank subject to a set of convex constraints \cite{wang1}, which can be formulated as
	\begin{equation}
		\hat{\mathbf{X}} = \arg \min_{\mathbf{X}} \;\mathrm{rank}(\mathbf{X}) \quad \mathrm{s.t.} \ \ f_i(\mathbf{X}) \le 0, \quad i=1, \ldots, m, 
		\label{Rank Func}
	\end{equation}
	where $\mathrm{rank}(\cdot)$ is the rank function, $f_i$ is a Lipschits convex function (usually the loss). However, directly minimizing the rank is NP hard and can not be solved in polynomial time. A widely used approach is to substitute the rank function with the nuclear norm, which can be formulated as 
	\begin{equation}
		\hat{\mathbf{X}} = \arg \min_{\mathbf{X}} \;\Vert \mathbf{X} \Vert_* \quad \mathrm{s.t.} \ \ f_i(\mathbf{X}) \le 0, \quad i=1, \ldots, m, 
		\label{Nuclear_cons}
	\end{equation}
	As proven by Fazel and Maryam \cite{tighest}, nuclear norm is the tighest convex relaxation of the original rank function. Moreover, Candès and Recht \cite{Candes} proved that the low rank matrix $\hat{\mathbf{X}}$ can be exactly recovered by nuclear norm minimization (NNM) under certain sampling conditions. Therefore, problem (\ref{Nuclear_cons}) can be solved by a flurry of nuclear norm-based algorithms, such as singular value thresholding \cite{SVT}, accelerated proximal gradient line search method \cite{APGL} and fixed point continuation with approximate singular value decomposition (SVD) \cite{FPCA}. However, nuclear norm tends to treat all singular values equally and over-shrink the large singular values. This is not very reasonable and might result in severe deviation from the desired solution. To alleviate this problem, a series of nonconvex low-rank regularizers have been studied, such as Schatten $p$-norm \cite{SNM}, $\ell_q$ norm with $0<q<1$ \cite{lq}, truncated $\ell_{1-2}$ norm \cite{TL12},  capped-$\ell_1$ norm \cite{capped_L1}, truncate nuclear norm \cite{TNNR}, and minimax concave penalty \cite{MCP}. Lots of works have demonstrated that nonconvex regularizers outperform nuclear norm both theoretically \cite{nonconvex_better_theo} and empirically \cite{nonconvex_better_emp}. In \cite{WNNM}, the famous weighted nuclear norm minimization (WNNM) model is proposed to solve grayscale image denoising problem. WNNM significantly improves the flexibility and capacity of the original NNM. It can achieve excellent denoising results while being highly efficient. However, WNNM use a fixed number of iterations to denoise all images. This is inflexible as various images contain different features and hence their best denoised version will be obtained after different numbers of iteration. Moreover, The performance of WNNM is not stable with the change of noise level. In order to enhance the stability and effectiveness of WNNM, Xie et al. propose the weighted Schatten $p$-norm minimization (WSNM) model \cite{WSNM}. WSNM generalizes WNNM and has more flexibility than WNNM. Moreover, WSNM outperforms WNNM under different noise levels. However, solving WSNM is time-consuming since it no longer has a closed-form solution as in the WNNM. Hence it has to be solved by the generalized iterated shrinkage algorithm.
	\par 
	Despite quantities of work on grayscale image denoising, color image denoising recieves less research attention in the past decade. Importantly, as the production and utilization of color images getting extremely popular nowadays, noise reduction for color images is bound to become an essential task in modern image processing system \cite{MNLF}. The most straightforward strategy is to extend the grayscale denoising methods to color images through a channel-wise manner. However, this strategy is inadequate to get satisfactory results since it fails to exploit the interchannel correlation between RGB components. Therefore, rational color denoising strategies should excavate the spectral interchannel correlation to achieve better performance. In \cite{two_color_strategies}, two practicable extension strategies are illustrated. The first strategy is to convert the color image from standard RGB (sRGB) space into a less correlated color space, such as YCbCr, and denoise each color channel independently. The groundbreaking work along this line is the color block-matching and 3D filtering (CBM3D) \cite{CBM3D}. CBM3D first converts the sRGB image into a luminance-chrominance space and applies BM3D to each channel separately. However, the color space conversion will complicate the structure of noise, and the interchannel correlation is not fully exploited by this strategy. The second strategy is to innovate coupling between RGB channels and design joint denoising algorithms \cite{MNLF}. In \cite{MNLF}, a rigorous penalize function is constructed to fully excavate the within and cross channel correlation of RGB components. Kong et al. \cite{sparRep4} and Lebrun et al. \cite{concatenate} concatenate color patches to vectors and denoise three channels simultaneously. Zhong et al. \cite{spatial_spectral} simultaneously model and use the spatial and spectral dependencies to denoise hyperspectral images. The multi-channel WNNM (MCWNNM) model \cite{MCWNNM} designs a diagnal form weight matrix to model the noise difference between channels based on the maximum a-posteior estimation. In addition to achieving state-of-the-art performance, MCWNNM also validates the joint denoising strategy outperforms other extension strategies. Nevertheless, MCWNNM still suffers from two main drawbacks. First, it is not capable of recovering the images with rich textures and details. Second, the optimal iteration number of MCWNNM for different images is not stable enough. To overcome these drawbacks, a multi-channel WSNM (MCWSNM) model is proposed in \cite{MCWSNM}. MCWSNM shows its superiority over MCWNNM with the rational setting of power $p$. However, MCWSNM has three main drawbacks. First of all, MCWSNM model can not obtain accurate analytical solution from endurable iterations in its iterative algorithm. This drawback constraints its performance critically. Moreover, solving MCWSNM is extremely expensive. In addition, MCWSNM is inadequate for competitiveness in reducing real noise.
	\par 
	To address the drawbacks mentioned above, in this paper we propose a new low-rank minimization model and utilize it to solve color image denoising problem. Specifically, we  utilize a sound nonconvex ``nuclear norm minus Frobenius norm'' (NNFN) regularizer \cite{NNFN}, and propose a multi-channel NNFN minimization (MC-NNFNM) model. The proposed model has three main advantages. First, our model satisfies adaptive shrinkage on singular values without assigning weights. Hence it can achieve satisfactory denoising performance without being hard to solve. Second, our model allows each variables to be updated with closed-form solutions in the alternating direction method of multipliers (ADMM) framework \cite{ADMM}. It makes our model bypass iterating and directly reach the desired solution in a single step. Third, the iteration numbers for different images to reach their best denoising version is stable. Furthermore, we design an efficient and accurate optimaization algorithm to solve the proposed MC-NNFNM model based on the ADMM framework. Meanwhile, we provide a theoretical guarantee to show each variable sequence generated by our algorithm converges to corresponding critical point. Extensive experiments demonstrate the effectiveness of our proposed model.
	\par 
	The rest of this paper is organized as follows. In Section 2 we describe the notations and briefly present the background of ADMM and several low-rank minimization methods. In Section 3 we formulate the problem, propose the MC-NNFNM model and give theoretical analyses. In Section 4 we report the experimental results. Finally, in Section 5 we conclude this paper.

	\section{Notations and Background}
	\subsection{Notations} Unless otherwise stated, lowercase boldface letters represent vectors and uppercase boldface letters represent matrices. $\mathbf{I}$ denotes the identity matrix. $(\cdot)^\top$ denotes the transpose operation. $tr(\cdot)$ is the trace of a matrix. 
	$\sigma(\mathbf{X})$ is the singular value vector of matrix $\mathbf{X}$ and $\sigma_i(\mathbf{X})$ is the $i$-th largest singular value. $Diag(\mathbf{x})$ is the diagnal matrix formed by vector $\mathbf{x}$. $\Vert \!\cdot\! \Vert_*$ is the nuclear norm, i.e., $\Vert \mathbf{X} \Vert_* = \sum_i \sigma_i(\mathbf{X})$. $\Vert \!\cdot\! \Vert_F$ is the Frobenius norm, i.e., $\Vert \mathbf{X} \Vert_F = \sqrt{tr(\mathbf{X}^\top \mathbf{X})}$. $\langle \!\cdot, \cdot\! \rangle$ stands for the inner product of two matrix, i.e., $\langle \mathbf{M}, \mathbf{N} \rangle = tr(\mathbf{M}^\top \mathbf{N})$. $\Vert \!\cdot\! \Vert_1$ is the vector $\ell_1$ norm, i.e., $\Vert \mathbf{x} \Vert_1 = \sum_i \vert x_i \vert$. $\Vert \!\cdot\! \Vert_2$ is the vector ${\ell_2}$ norm, i.e.,  $\Vert \mathbf{x} \Vert_2 = \sqrt{\sum_i  \vert x_i \vert^2}$.
	
	\subsection{Alternating Direction Method of Multipliers}
	ADMM is an effective and flexible tool to solve convex and nonconvex optimization problems. Over the decades, ADMM has been widely studied, re-invented and applied in different fields. The basic idea of ADMM is to decompose the original problem into a set of subproblems and solve them alternately. Consider optimization problems of the form
	\begin{equation}
		\min_{\mathbf{x}, \mathbf{z}} \; f(\mathbf{x}) + g(\mathbf{z}) \quad \mathrm{s.t.} \ \ \mathbf{Ax} + \mathbf{Bz} = \mathbf{c},
		\label{ADMM_0}
	\end{equation}
	where $f$ and $g$ are convex functions, $\mathbf{x} \in \mathbb{R}^n$, $\mathbf{z} \in \mathbb{R}^m$, $\mathbf{A} \in \mathbb{R}^{p \times n}$, $\mathbf{B} \in \mathbb{R}^{p \times m}$, and  $\mathbf{c} \in \mathbb{R}^p$. The augmented Lagrangian of (\ref{ADMM_0})
	\begin{equation}
		\mathcal{L}_{\rho} (\mathbf{x}, \mathbf{z}, \mathbf{y}) = f(\mathbf{x}) + g(\mathbf{z}) + \mathbf{y}^\top (\mathbf{Ax} + \mathbf{Bz} - \mathbf{c}) + \frac{\rho}{2} \Vert \mathbf{Ax} + \mathbf{Bz} - \mathbf{c} \Vert_2^2,
	\end{equation}
	where $\mathbf{y}$ is called the dual variable or Lagrange multiplier, and $\rho>0$ is the penalty parameter. At the $k$-th iteration, ADMM updates $\mathbf{x}$ and $\mathbf{z}$ alternately by minimizing the augmented Lagrangian $\mathcal{L} (\mathbf{x}, \mathbf{z}, \mathbf{y})$.
    	\begin{numcases}{}
    	    \mathbf{x}^{k+1} = \arg \min_{\mathbf{x}}\limits \ \mathcal{L} (\mathbf{x}, \mathbf{z}^k, \mathbf{y}^k), \\
    	    \mathbf{z}^{k+1} = \arg \min_{\mathbf{z}}\limits \ \mathcal{L} (\mathbf{x}^{k+1}, \mathbf{z}, \mathbf{y}^k).
    	\end{numcases}
	Then ADMM updates the Lagrange multiplier
	\begin{equation}
		\mathbf{y}^{k+1} = \mathbf{y}^k + \rho (\mathbf{Ax}^{k+1} + \mathbf{Bz}^{k+1} - \mathbf{c}).
	\end{equation}
	For convex problems, the convergence of ADMM has strong theoretical guarantees \cite{convex_conv1, convex_conv3}, which demonstrate ADMM can converge to the optimal point under mild conditions. While for nonconvex problems, ADMM can only converge under certain restrictive conditions \cite{noncon_conv2}. However, ADMM still obtains good preformance on many nonconvex problems, such as matrix completion \cite{ADMM_app_nonc1} and matrix-tensor factorization \cite{ADMM_app_nonc2}. In this paper we utilize ADMM to solve the nonconvex optimization problem in our model.
	
	\subsection{Existing Low-Rank Minimization Methods for Image Denoising}
	Color image denoising based on low-rank minimization has led to several state-of-the-art methods in the past decade. In \cite{NSP}, the nonlocal spectral prior (NSP) is established, which makes it reasonable to use the low-rank minimization method to design denoising algorithms. The NSP accounts for the fact that there are many similar patterns across a natrual image, and hence the matrix formed by nonlocal similar patches should be low rank with its singular values being heavy-tailed. Based on NSP, low-rank minimization is allowed to be performed on each patch matrix to estimate its clean low rank version. The clean image can be generated using all denoised patch matrices.
	\par 
	The groundbreaking work in this field is achieved by the WNNM model \cite{WNNM}, which is described as
	\begin{equation}
		\min_{\mathbf{X}} \Vert \mathbf{Y} - \mathbf{X} \Vert_F^2 + \lambda \Vert \mathbf{X} \Vert_{\mathbf{w},*},
	\end{equation}
	where $\Vert \mathbf{X} \Vert_{\mathbf{w},*} = \sum_i w_i \sigma_i(\mathbf{X})$ is the weighted nuclear norm of matrix $\mathbf{X}$, and $\mathbf{w} = [w_1, \ldots, w_n]^\top$ is the non-negative weight vector. WNNM assigns weithts on different sigular values to alleviate the biasd problem of the original NNM. The global optimal solution of WNNM can be reached in closed-form when the weights satisfy a non-descending order, i.e. $0 \le w_1 \le \ldots \le w_n$. Interestingly, the non-descending weights just agree with the physical meaning of singular values and make WNNM satisfy the adaptive shrinkage property. Hence WNNM can achieve state-of-the-art performance with high efficiency. In \cite{WSNM}, Xie et al. generalize the WNNM and propose the weighted Schatten $p$-norm minimization (WSNM) model to solve grayscale image denoising problem. Under non-descending weight permutation, WSNM can be decomposed into independent $\ell_p$-norm subproblems and be solved by the generalized soft-thresholding (GST) \cite{GST} algorithm. Therefore, WSNM is less efficient than WNNM. However, WSNM outperforms WNNM not only on denoising results but also on the stability for different noise levels.
	\par 
	The key issue for designing feasible color image denoising methods lies in modeling and utilizing the interchannel correlation of RGB components and the cross-channel difference of color noise. The multi-channel WNNM (MCWNNM) model \cite{MCWNNM}, which is described as (\ref{MCWNNM}), introduces a weight matrix to balance the contributions of RGB channels based on their noise strength. 
	\begin{equation}
		\min_{\mathbf{X}} \Vert \mathbf{W} (\mathbf{Y} - \mathbf{X}) \Vert_F^2 + \Vert \mathbf{X} \Vert_{\mathbf{w}, *}.
		\label{MCWNNM}
	\end{equation}
	However, MCWNNM no longer has a global optimum with closed form solution because of the weight matrix $\mathbf{W}$. Therefore, ADMM is resorted to solve problem (\ref{MCWNNM}). Due to the non-convexity of MCWNNM model, an inflated and unbounded penalty parameters $\rho_k$ is used for each iteration, i.e. $\rho_k \rightarrow +\infty$ as $k \rightarrow \infty$, to make ADMM converge. In practice, modestly accurate results produced by ADMM within several iterations are adopted. And the accuracy is sufficient for the color image denoising problem. MCWNNM has multiple contributions. First, it achieves state-of-the-art performance on color image denoising. Besides, it first introduces a weight matrix to model and use the cross-channel difference of noise. Meanwhile, it validates that the joint denoising stategy outperforms other strategies for color image denoising. The multi-channel WSNM (MCWSNM) model \cite{MCWSNM}, which can be expressed as (\ref{MCWSNM}), utilizes the same weight matrix to jointly process three channels and simultaneously consider their noise differences.
	\begin{equation}
		\min_{\mathbf{X}} \Vert \mathbf{W} (\mathbf{Y} - \mathbf{X}) \Vert_F^2 + \Vert \mathbf{X} \Vert_{\mathbf{w}, S_p}^{p},
		\label{MCWSNM}
	\end{equation}
	where $\Vert \mathbf{X} \Vert_{\mathbf{w}, S_p}^{p} = \sum_i w_i \sigma_i^p$ is the weight Schatten $p$-norm. With ADMM, MCWSNM can be decomposed into two subproblems, and one of them can be solved by GST. MCWSNM extends the benchmark WSNM to color image denoising and achieves competitive performance.
	
	\begin{figure}[tb]
		\centering
		\includegraphics[width=16cm]{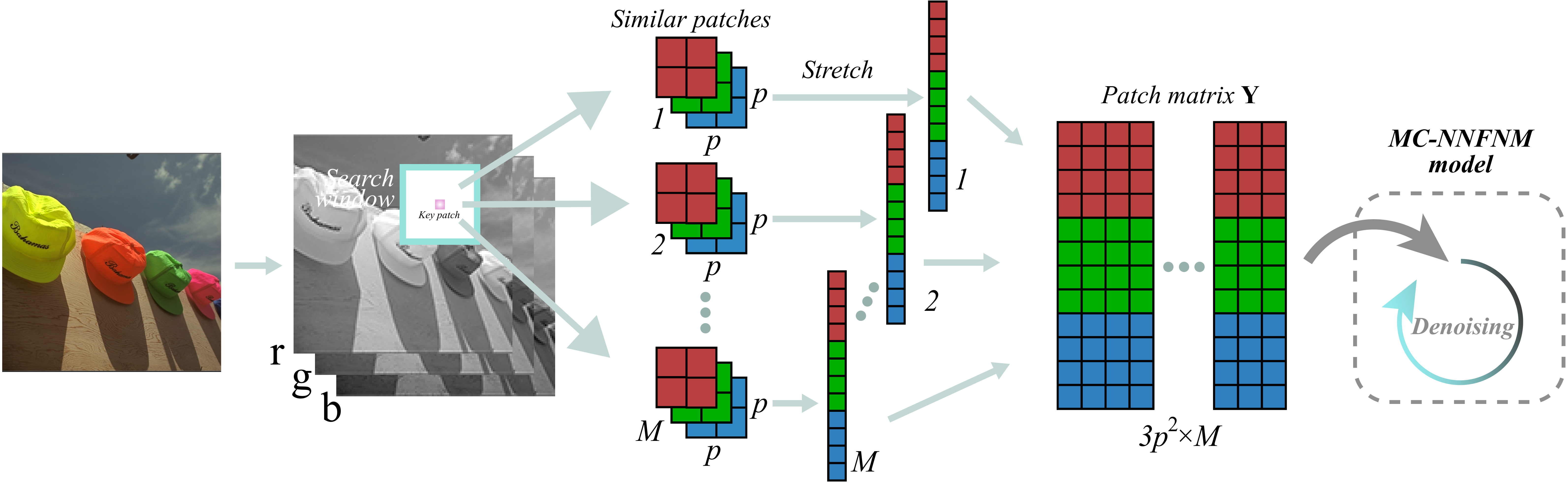}
		\caption{\centering The process of generating a patch matrix from a key patch.}
		\label{form_patch_matrix}
	\end{figure}
	
\section{The Proposed Model}
\subsection{Problem Formulation}
Image denoising aims to estimate the underlying clean image $\mathbf{x}_c \in \mathbb{R}^{m \times n}$ from its noisy observation 
\begin{equation}
	\mathbf{y}_c = \mathbf{x}_c + \mathbf{n}_c, 
\end{equation}
where $c \in \lbrace r,g,b \rbrace$ is the index of R, G, B channels and $\mathbf{n}_c$ is the noise in channel $c$. In past decade, low-rank minimization methods, which exploits the the low-rank property of the redundant nonlocal similar patches over an image, has shown remarkable denoising performance. Given a noisy color image $\mathbf{y}_c$, we assign a number of key patches of size $p \times p \times 3$ across the image with a fixed interval. For each key patch, we find its $M$ most similar patches (including itself) in a search window around it. For each extracted similar patch of size $p \times p \times 3$, we stretch it to a column vector $\mathbf{y} = [ \mathbf{y}_r^\top, \mathbf{y}_g^\top, \mathbf{y}_b^\top ]^\top$, where $\mathbf{y}_r, \mathbf{y}_g, \mathbf{y}_b \in \mathbb{R}^{p^2}$ are the corresponding patch vectors of R, G, B channels. Then we stack the $M$ vectors column by column to form a noisy patch matrix $\mathbf{Y} = \mathbf{X} + \mathbf{N} \in \mathbb{R}^{3p^2 \times M}$, where $\mathbf{X}$ is the underlying clean matrix and  $\mathbf{N}$ is the noise matrix. \figref{form_patch_matrix} demonstrates the procedure of generating a patch matrix $\mathbf{Y}$ from a key patch. Low-rank minimization model is performed on $\mathbf{Y}$ to estimate the clean patch matrix $\mathbf{X}$. The clean image can be generated from all denoised patch matrices.
\par 
Our proposed multi-channel “nuclear norm minus Frobenius norm” minimization (MC-NNFNM) model aims to find a matrix $\mathbf{X}$ as close to the observation $\mathbf{Y}$ as possible under the Frobenius norm data fidelity and the NNFN regularization:
\begin{equation}
	\hat{\mathbf{X}} = \arg \min_{\mathbf{X}} \Vert \mathbf{W}(\mathbf{Y} - \mathbf{X}) \Vert_F^2 + \lambda (\Vert \mathbf{X} \Vert_* - \alpha \Vert \mathbf{X} \Vert_F ),
	\label{NNFNM}
\end{equation}
where $\mathbf{W}$ is a weight matrix to model the noise cross-channel difference, $\lambda$ is a trade-off parameter to balance the two terms, and $\alpha$ is a non-negative parameter. Using a nonconvex NNFN regularizer with several virtues and sound theoretical guarantees \cite{NNFN}, the proposed MC-NNFNM model can shrink singular values adaptively without assigning weights on them. Therefore, our model can achieve satisfactory denoising results while avoid being hard to solve. The solution of our model is presented in detail in next section. Before that, we discuss the determination of the weight matrix $\mathbf{W}$.
\par 
We determine $\mathbf{W}$ based on the maximum a posteriori (MAP) estimation. Given the observation $\mathbf{Y} \in \mathbb{R}^{3p^2 \times M}$, the MAP estimate of matrix $\mathbf{X}$ can be obtained by
\begin{align}
	\hat{\mathbf{X}} =  \arg \max_{\mathbf{X}} \ln P(\mathbf{X}|\mathbf{Y}) 
	= \arg \max_{\mathbf{X}} \ln \frac{P(\mathbf{Y} | \mathbf{X}) P(\mathbf{X})}{P(\mathbf{Y})} = \arg \max_{\mathbf{X}} \{\ln P(\mathbf{Y} | \mathbf{X}) + \ln P(\mathbf{X}) \}.
	\label{map}
\end{align}
According to \cite{Assume_AWGN}, we assume the noise in each channel is independently and identically distributed with Gaussian distribution and standard deviations $\{ \sigma_r, \sigma_g, \sigma_b \}$. Thus the likelihood term $P(\mathbf{Y} | \mathbf{X})$ can be formulated as 
\begin{equation}
	P(\mathbf{Y} | \mathbf{X}) =  \prod_{c \in \{r, g, b\}} (\frac{1}{\sqrt{2\pi} \sigma_c } )^{M} \
	\exp (-\frac{1}{2\sigma_c^2} \Vert \mathbf{Y}_c - \mathbf{X}_c \Vert_F^2 ).
\end{equation}
Since the minimum NNFN property is imposed on the underlying matrix $\mathbf{X}$, we let $P(\mathbf{X})$ follow
\begin{equation}
	P(\mathbf{X}) \propto \exp \left(-\lambda (\Vert \mathbf{X} \Vert_* - \alpha \Vert \mathbf{X} \Vert_F ) \right).
\end{equation}
Therefore, problem (\ref{map}) can be rewritten as
\begin{align}
	\hat{\mathbf{X}} &= \arg \min_{\mathbf{X}} \sum_{c \in \{r, g, b\}} \frac{1}{\sigma_c^2} \Vert \mathbf{Y}_c - \mathbf{X}_c \Vert_F^2 + \lambda (\Vert \mathbf{X} \Vert_* - \alpha \Vert \mathbf{X} \Vert_F ) \notag \\
	&= \arg \min_{\mathbf{X}} \Vert \mathbf{W}(\mathbf{Y} - \mathbf{X}) \Vert_F^2 + \lambda (\Vert \mathbf{X} \Vert_* - \alpha \Vert \mathbf{X} \Vert_F ),
\end{align}
with 
\begin{equation}
	\mathbf{W} = 
	\begin{pmatrix}
		\sigma_r^{-1} \mathbf{I} & \mathbf{0} & \mathbf{0} \\
		\mathbf{0} & \sigma_g^{-1} \mathbf{I} & \mathbf{0} \\
		\mathbf{0} & \mathbf{0} & \sigma_b^{-1} \mathbf{I}
	\end{pmatrix},
\end{equation}
where $\mathbf{I} \in \mathbb{R}^{p^2 \times p^2}$ is the identity matrix. $\mathbf{W}$ is determined by the noise standard deviation in three channels. It can model the noise difference and allow our proposed model to perform joint denoising.

\subsection{Optimization}
	We propose an efficient and accurate algorithm to solve the proposed MC-NNFNM model based on the alternating direction method of mulitpliers (ADMM) framework. First, we rewrite model (\ref{NNFNM}) as follows:
	\begin{equation}
		\min_{\mathbf{X,Z}} \left\Vert \mathbf{W}\left( \mathbf{X}- \mathbf{Y} \right) \right\Vert_F^2 + \lambda \left( \left\Vert \mathbf{Z} \right\Vert_* - \alpha \left\Vert \mathbf{Z} \right\Vert_F \right),\quad
		\mathrm{ s.t. } \ \ \mathbf{X} = \mathbf{Z}.
		\label{NNFNM2}
	\end{equation}
	The augmented Lagrangian of optimization problem (\ref{NNFNM2}) can be formulated as
	\begin{equation}
		\mathcal{L}_\rho \left( \mathbf{X}, \mathbf{Z}, \mathbf{A} \right) = \left \Vert \mathbf{W}\left( \mathbf{Y}-\mathbf{X} \right) \right\Vert_F^2 + \lambda \left( \left\Vert \mathbf{Z} \right\Vert_* - \alpha \left\Vert \mathbf{Z} \right\Vert_F \right) + \langle \mathbf{A}, \mathbf{X}-\mathbf{Z} \rangle + \frac{\rho}{2} \left\Vert \mathbf{X} - \mathbf{Z} \right\Vert_F^2,
	\end{equation}
	where $\mathbf{A}$ is the augmented Lagrange multiplier and $\rho > 0$ is the penality parameter. Our proposed algorithm updates $\mathbf{X}$ and $\mathbf{Z}$ alternately by minimizing the augmented Lagrangian $\mathcal{L}_{\rho} (\mathbf{X}, \mathbf{Z}, \mathbf{A})$. Then update $\mathbf{A}$ and $\rho$. Specifically, problem (\ref{NNFNM2}) can be solved via the following four steps.
	\begin{enumerate}[fullwidth,itemindent=0em,label=(\bf \arabic*) ]
		\item \textbf{$\mathbf{X}$-update}: Fix $\mathbf{Z}_k$, $\mathbf{A}_k$ and $\rho_k$, minimize $\mathcal{L}_{\rho_k} (\mathbf{X}, \mathbf{Z}_k, \mathbf{A}_k)$ for $\mathbf{X}_{k+1}$ as follows:
	\begin{align}
		\mathbf{X}_{k+1} &= \arg \min_{\mathbf{X}} \mathcal{L}_{\rho_k} (\mathbf{X}, \mathbf{Z}_k, \mathbf{A}_k) \notag \\
		&= \arg \min_{\mathbf{X}} \Vert \mathbf{W}\left( \mathbf{Y}-\mathbf{X} \right) \Vert_F^2 + \lambda \left( \left\Vert \mathbf{Z}_k \right\Vert_* - \alpha \left\Vert \mathbf{Z}_k \right\Vert_F \right) + \langle \mathbf{A}_k, \mathbf{X} - \mathbf{Z}_k \rangle + \frac{\rho_k}{2} \left\Vert \mathbf{X} - \mathbf{Z}_k \right\Vert_F^2.
		\label{upX}
	\end{align} 
	Ignoring constant terms, problem (\ref{upX}) can be rewritten as
	\begin{equation}
		\mathbf{X}_{k+1} = \arg \min_{\mathbf{X}} \Vert \mathbf{W} ( \mathbf{Y}-\mathbf{X}) \Vert_F^2 + \frac{\rho_k}{2} \Vert \mathbf{X} - \mathbf{Z}_k + \rho_k^{-1} \mathbf{A}_k \Vert_F^2,
		\label{upX2}
	\end{equation}
	Problem (\ref{upX2}) is convex, and has a closed-form solution:
	\begin{equation}
		\mathbf{X}_{k+1} = ( \mathbf{W}^\top \mathbf{W} + \frac{\rho_k}{2} \mathbf{I} )^{-1} ( \mathbf{W}^\top \mathbf{WY} + \frac{\rho_k}{2}\mathbf{Z}_k - \frac{1}{2}\mathbf{A}_k ).
		\label{up_X_close}
	\end{equation}
	The dominate cost of equation (\ref{up_X_close}) lies in calculating $\mathbf{W}^\top \mathbf{W} \mathbf{Y}$. Thus updating $\mathbf{X}$ costs $\mathcal{O} (p^4M)$.
	
	\item \textbf{$\mathbf{Z}$-update}: Fix $\mathbf{X}_{k+1}$, $\mathbf{A}_k$ and $\rho_k$, minimize $\mathcal{L}_{\rho_k} (\mathbf{X}_{k+1}, \mathbf{Z}, \mathbf{A}_k)$ for $\mathbf{Z}_{k+1}$ as follows:
	\begin{align}
		\mathbf{Z}_{k+1} &= \arg \min_{\mathbf{Z}} \mathcal{L}_{\rho_k} (\mathbf{X}_{k+1}, \mathbf{Z}, \mathbf{A}_k) \notag \\
		&= \frac{\rho_k}{2} \left\Vert \mathbf{Z} - \left( \mathbf{X}_{k+1} + \rho_k^{-1}\mathbf{A}_k \right) \right\Vert_F^2  + \lambda \left( \Vert\mathbf{Z}\Vert_* - \alpha \Vert\mathbf{Z}\Vert_F \right).
		\label{upZ}
	\end{align}
	Optimization problem (\ref{upZ}) can be solved under the $L_1 - \alpha L_2$ minimization framework \cite{L12}. Let $ \mathbf{X}_{k+1} + \rho_k^{-1}\mathbf{A}_k = \mathbf{U}_k Diag([\sigma_1, \sigma_2, \dots, \sigma_M]^\top) \mathbf{V}_k^\top $ be its SVD. According to the Proposition 2 in \cite{NNFN}, the global optimum of (\ref{upZ}) has the form 
	\begin{equation}
		\mathbf{Z}_{k+1} = \mathbf{U}_k Diag \left(\mathbf{prox}_{\frac{\lambda}{\rho_k}\Vert \cdot \Vert_{1-\alpha2} } \left( [ \sigma_1, \sigma_2, \dots, \sigma_M ]^\top \right) \right) \mathbf{V}_k^\top,
		\label{upZ2}
	\end{equation}
	where $\mathbf{prox}_{\lambda/\rho_k \Vert \cdot \Vert_{1-\alpha2} }$ is the proximal operator for $L_1 - \alpha L_2$ minimization. $\mathbf{prox}_{\lambda/\rho_k \Vert \cdot \Vert_{1-\alpha2} }$ returns a vector, denoted as $[\hat{\sigma}_1, \ldots, \hat{\sigma}_M]^\top$, and has closed-form solution $( i=1, 2, \dots, M )$:
	\begin{equation}
		\hat{\sigma}_i = \left\{ 
		\begin{aligned}
			& \frac{\Vert z \Vert_2 + \alpha \lambda/\rho_k}{\Vert z \Vert_2} z,  &  &\quad \sigma_1 \ge \lambda/\rho_k, \\
			& 0, &  &\quad \sigma_1 < \lambda/\rho_k,
		\end{aligned}
		\right. 
	\end{equation}
	where $z$ is the soft shrinkage, i.e. $z_i = \max \lbrace \sigma_i - \lambda / \rho_k, 0 \rbrace$. To sum up, $\mathbf{Z}_{k+1}$ can be updated with
	\begin{equation}
		\mathbf{Z}_{k+1} = \mathbf{U}_k Diag\left(\mathbf{prox}_{\frac{\lambda}{\rho_k} \Vert \cdot \Vert_{1-\alpha2} } \left(\sigma(\mathbf{X}_{k+1} + \rho^{-1} \mathbf{A}_k )\right)\right) \mathbf{V}_k^\top.
	\end{equation}
	The cost for updating $\mathbf{Z}$, which is dominated by the SVD, is $\mathcal{O} (p^4M + p^2M^2)$.
	
	\item $\mathbf{A}$\textbf{-update}: Fix $\mathbf{X}_{k+1}$, $\mathbf{Z}_{k+1}$ and $\rho$, calculate $\mathbf{A}_{k+1}$ as follows:
	\begin{equation}
		\mathbf{A}_{k+1} = \mathbf{A}_{k} + \rho_k \left(\mathbf{X}_{k+1} - \mathbf{Z}_{k+1} \right).
		\label{upA}
	\end{equation}
	This step costs $\mathcal{O} (p^2M)$.
	
	\item $\rho$\textbf{-update}: Fix $\mathbf{X}_{k+1}$, $\mathbf{Z}_{k+1}$ and $\mathbf{A}_{k+1}$, update $\rho_{k+1}$ as:
	\begin{equation}
		\rho_{k+1} = \mu\rho_{k},
	\end{equation}
	where $\mu > 1$. This step aims to make the sequence $\lbrace \rho_k \rbrace$ unbounded, which is crucial to guarantee our algorithm converges. 
\end{enumerate}
\par
\begin{figure}[tb]
	\centering
	\includegraphics[width=8cm]{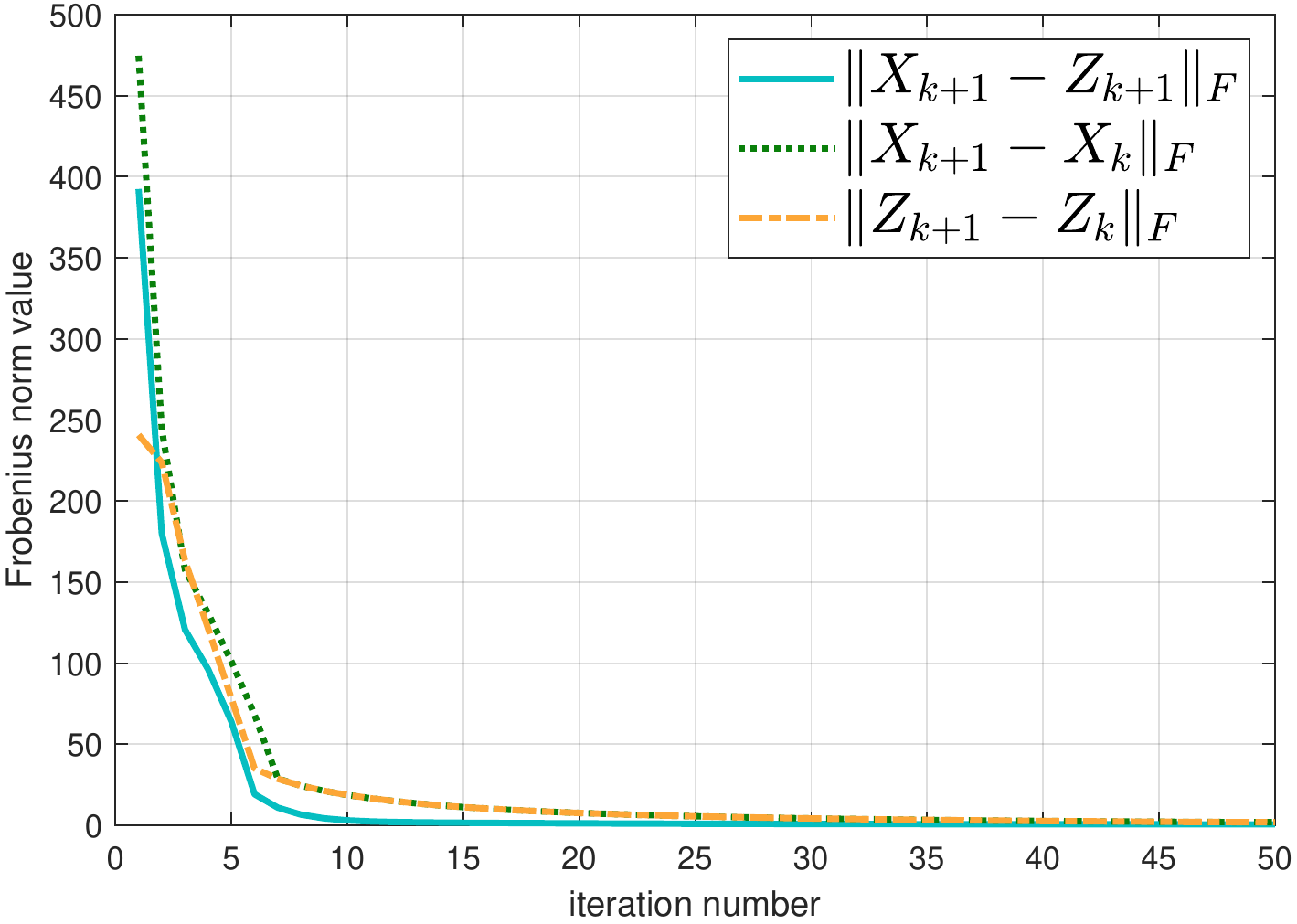}
	\caption{The convergence curves of $\Vert \mathbf{X}_{k+1} - \mathbf{Z}_{k+1} \Vert_F$, $\Vert \mathbf{X}_{k+1} - \mathbf{X}_k \Vert_F$ and $\Vert \mathbf{Z}_{k+1} - \mathbf{Z}_k \Vert_F$ of image ``kodim01'' in Kodak PhotoCD dataset. }
	\label{3converge}
\end{figure}

The above updating steps are repeated sequentially until the stopping criterion is satisfied or the number of iteration exceeds a threshold set beforehand. The stopping criterion involves a simultaneous satisfaction on
\begin{align}
	(a)\; \Vert \mathbf{X}_{k+1} - \mathbf{Z}_{k+1} \Vert_F \le \tau, \quad
	(b)\; \Vert \mathbf{X}_{k+1} - \mathbf{X}_{k} \Vert_F \le \tau, \quad 
	(c)\; \Vert \mathbf{Z}_{k+1} - \mathbf{Z}_{k} \Vert_F \le \tau, \notag
\end{align}
where $\tau > 0$ is a small tolerance value. The optimization algorithm is summarized in Algorithm \ref{admm}. We provide \figref{3converge} to intuitively show that $\Vert \mathbf{X}_{k+1} - \mathbf{Z}_{k+1} \Vert_F$, $\Vert \mathbf{X}_{k+1} - \mathbf{X}_{k} \Vert_F$ and $\Vert \mathbf{Z}_{k+1} - \mathbf{Z}_{k} \Vert_F$ do tend to zero simultaneously during the iteration process of Algorithm \ref{admm}. And the rationality of the stopping criterion will be proved in detail in section \textit{3.4}.

\begin{algorithm}[tb] 
	\caption{Solve MC-NNFNM via ADMM}
	\label{admm}
	\textbf{Input}: Noisy matrix $\mathbf{Y}$, weight matrix $\mathbf{W}$, $\mu>1$, $\tau>0$, iteration threshold $K_1$;\\
	\textbf{Initialization}: $\mathbf{X}_0 = \mathbf{Z}_0 = \mathbf{A}_0 = \mathbf{0}$, $\rho_0 > 0$, $\mathrm{T = False}$, $k = 0$;
	\begin{algorithmic}[1] 
		\WHILE{$\mathrm{T} == \mathrm{False}$}
		\STATE Update $\mathbf{X}$ by (\ref{up_X_close});
		\STATE Update $\mathbf{Z}$ by solving the problem (\ref{upZ});
		\STATE Update $\mathbf{Z}$ by (\ref{upA});
		\STATE Update $\rho$: $\rho_{k+1} = \mu\rho_k$;
		\STATE $k \leftarrow k+1$
		\IF {(Stopping criterion is satisfied) or ($k \le K_1$)}
		\STATE $\mathrm{T} \leftarrow \mathrm{True}$
		\ENDIF
		\ENDWHILE
		\STATE \textbf{return} $\mathbf{Z}$
	\end{algorithmic}
\end{algorithm}

\begin{algorithm}[tb] 
	\caption{Color image denoising by MC-NNFNM}
	\label{framework}
	\textbf{Input}: Noisy image $\mathbf{y}_c$, noise levels $(\sigma_r, \sigma_g, \sigma_b)$, $K_2$;\\
	\textbf{Initialization}: $\hat{\mathbf{x}}_c^{(0)} = \mathbf{y}_c, \mathbf{y}_c^{(0)} = \mathbf{y}_c$;
	\begin{algorithmic}[1] 
		\STATE Set $\mathbf{y}_c^{(k)} = \mathbf{x}_c^{(k-1)}$;
		\FOR{$k = 1: K_2$}
		\STATE Extract $N$ key patches $\lbrace \mathbf{y}_j \rbrace_{j=1}^{N}$ from $\mathbf{y}_c^{(k)}$;
		\FOR {each patch $\mathbf{y}_j$}
		\STATE Search its $M$ similar patches to form $\mathbf{Y}_j$;
		\STATE Perform the MC-NNFNM model (\ref{NNFNM2}) on $\mathbf{Y}_j$ to obtain the estimated $\mathbf{X}_j$;
		\ENDFOR
		\ENDFOR
		\STATE Aggregate $\lbrace \mathbf{X}_j \rbrace_{j=1}^{N}$ to form the estimated $\hat{\mathbf{x}}_c^{(k)}$;
		\STATE \textbf{return} $\hat{\mathbf{x}}_c^{(K_2)}$
	\end{algorithmic}
\end{algorithm}

\subsection{The Denoising Scheme}
	Given a noisy image $\mathbf{y}_c$, we first extract $N$ key patches $\lbrace \mathbf{y}_j \rbrace_{j=1}^{N}$ across the image with a fixed interval. For each key patch $\mathbf{y}_j$ of size $p \times p \times 3$, we search its $M$ similar patches (including itself) in a search window around it of size $S \times S$. The $M$ patches are to form a noisy patch matrix $\mathbf{Y}_j \in \mathbb{R}^{3p^2 \times M}$. The MC-NNFNM model (\ref{NNFNM2}) is performed on each patch matrix $\mathbf{Y}_j$ to estimate its noise-free version $\mathbf{X}_j$. The noise-free image $\hat{\mathbf{x}}_c$ can be yield using all denoised patch matrices $\lbrace \mathbf{X}_j \rbrace_{j=1}^{N}$. To obtain better denoising results, above procedures are executed several times. The whole scheme for color image denoising is summarized in Algorithm \ref{framework}.

\subsection{Algorithm Analysis}
	In this section we analyze the convergence and complexity of Algorithm \ref{admm}. Due to the non-convexity of MC-NNFNM model, the convergence of Algorithm \ref{admm} is difficult to strictly analyze. Thus we present a weak convergence result in Theorem \ref{Th1} as a succinct construction of a rational stopping criterion.
	\begin{theorem}
		Assume that the sequence of parameter $\left\{\rho_k\right\}$ is unbounded. Then the sequences $\left\{\mathbf{X}_k\right\}$, $\left\{\mathbf{Z}_k\right\}$ and $\left\{\mathbf{A}_k\right\}$ in Algorithm \ref{admm} satisfy:
		\begin{align}
			(a) \lim_{k \rightarrow \infty} \Vert \mathbf{X}_{k+1} - \mathbf{Z}_{k+1} \Vert_F = 0; \quad (b) \lim_{k \rightarrow \infty} \Vert \mathbf{X}_{k+1} - \mathbf{X}_k \Vert_F = 0; \quad (c) \lim_{k \rightarrow \infty} \Vert \mathbf{Z}_{k+1} - \mathbf{Z}_k \Vert_F = 0. \notag
		\end{align} \label{Th1}
	\end{theorem}
	\noindent \textit{Proof.} 1. We first prove that the sequence of the augmented Lagrangian multiplier $\left\{\mathbf{A}_k\right\}$ is upper bounded.
	\begin{align}
		\Vert \mathbf{A}_{k+1} \Vert_F^2 &= \Vert \mathbf{A}_k + \rho_k \left(\mathbf{X}_{k+1} - \mathbf{Z}_{k+1} \right) \Vert_F^2 \notag \\ 
		&= \rho_k^2 \left\Vert \left(\rho_k^{-1} \mathbf{A}_k + \mathbf{X}_{k+1} \right) - \mathbf{Z}_{k+1} \right\Vert_F^2 \notag \\
		&= \rho_k^2 \Vert \mathbf{U}_k Diag([\sigma_1, \ldots, \sigma_M]^\top) \mathbf{V}_k^\top - \mathbf{U}_k Diag([\hat{\sigma}_1, \ldots, \hat{\sigma}_M]^\top) \mathbf{V}_k^\top \Vert_F^2 \notag \\
		&= \rho_k^2 \sum_{i=1}^{M} \left(\sigma_i - \hat{\sigma}_i \right)^2 \ \le \rho_k^2 \sum_{i=1}^{M} (\lambda/\rho_k)^2 \ = \lambda^2 M. \notag
	\end{align}
	The inequality in the last step holds. If $\sigma_i \ge \lambda/\rho_k$, then $\hat{\sigma}_i = \frac{\Vert z \Vert_2 + \lambda/ \rho_k}{\Vert z \Vert_2} \left(\sigma_i - \lambda/\rho_k \right)$. We can deduce $\sigma_i - \hat{\sigma}_i \le \lambda/\rho_k$. If $\sigma_i < \lambda/\rho_k$, then $\hat{\sigma}_i = 0$. We have $\sigma_i - \hat{\sigma}_i = \sigma_i < \lambda/\rho_k$.
	\par 
	2. We then prove the augmented Lagrangian sequence $\left\{ \mathcal{L}_{\rho_k} \left( \mathbf{X}_{k+1}, \mathbf{Z}_{k+1}, \mathbf{A}_k \right) \right\}$ is upper bounded. 
	Since $\mathbf{A}_{k+1} = \mathbf{A}_{k} + \rho_k \left( \mathbf{X}_{k+1} - \mathbf{Z}_{k+1} \right)$, we have
	\begin{align}
		&\mathcal{L}_{\rho_{k+1}} \left( \mathbf{X}_{k+1}, \mathbf{Z}_{k+1}, \mathbf{A}_{k+1} \right) \notag \\
		&= \mathcal{L}_{\rho_k} \left( \mathbf{X}_{k+1}, \mathbf{Z}_{k+1}, \mathbf{A}_{k}, \right) + \left \langle \mathbf{A}_{k+1}-\mathbf{A}_{k}, \mathbf{X}_{k+1} - \mathbf{Z}_{k+1} \right \rangle + (\rho_{k+1} - \rho_k)/2 \left\Vert \mathbf{X}_{k+1} - \mathbf{Z}_{k+1} \right\Vert_F^2 \notag \\
		&= \mathcal{L}_{\rho_k} \left( \mathbf{X}_{k+1}, \mathbf{Z}_{k+1}, \mathbf{A}_{k}, \right) + \left \langle \mathbf{A}_{k+1}-\mathbf{A}_{k}, (\mathbf{A}_{k+1} - \mathbf{A}_{k})/\rho_k \right \rangle + (\rho_{k+1} - \rho_k)/2 \left \Vert (\mathbf{A}_{k+1} - \mathbf{A}_{k})/\rho_k \right \Vert_F^2 \notag \\
		&= \mathcal{L}_{\rho_k} \left( \mathbf{X}_{k+1}, \mathbf{Z}_{k+1}, \mathbf{A}_{k}, \right) + \frac{\rho_{k+1} + \rho_k}{2\rho_k^2} \Vert \mathbf{A}_{k+1} - \mathbf{A}_k \Vert_F^2. \notag
	\end{align}
	Since $\left\{ \mathbf{A}_k \right\}$ is upper bounded, the sequence $\left\{ \mathbf{A}_{k+1} - \mathbf{A}_k \right\}$ is also upper bounded. Denoting the upper bound as $a$ (i.e. $\forall k \ge 0,\ \Vert \mathbf{A}_{k+1} - \mathbf{A}_k \Vert_F \le a $), we have
	\begin{align}
		\mathcal{L}_{\rho_{k+1}} \left( \mathbf{X}_{k+1}, \mathbf{Z}_{k+1}, \mathbf{A}_{k+1} \right) &\le \mathcal{L}_{\rho_k} \left( \mathbf{X}_{k+1}, \mathbf{Z}_{k+1}, \mathbf{A}_{k} \right) + \frac{\rho_{k+1} + \rho_k}{2\rho_k^2}a^2 \notag \\
		&\le \mathcal{L}_{\rho_0} \left( \mathbf{X}_{1}, \mathbf{Z}_{1}, \mathbf{A}_{0} \right) + a^2{\sum}_{k=0}^{\infty}\frac{\rho_{k+1} + \rho_k}{2\rho_k^2} \notag \\
		&= \mathcal{L}_{\rho_0} \left( \mathbf{X}_{1}, \mathbf{Z}_{1}, \mathbf{A}_{0} \right) + a^2{\sum}_{k=0}^{\infty}\frac{\mu + 1}{2\mu^k\rho_0} \notag \\
		&\le \mathcal{L}_{\rho_0} \left( \mathbf{X}_{1}, \mathbf{Z}_{1}, \mathbf{A}_{0} \right) + \frac{a^2}{\rho_0}{\sum}_{k=0}^{\infty}\frac{1}{\mu^{k-1}}. \notag
	\end{align}
	The last inequality holds since $\mu>1$ and $\mu+1<2\mu$. As $\sum_{k=0}^{\infty}\frac{1}{\mu^{k-1}} < \infty$, the sequence $\lbrace \mathcal{L} \left( \mathbf{X}_{k+1}, \mathbf{Z}_{k+1}, \mathbf{A}_k, \rho_k \right) \rbrace$ is upper bounded.
	\par 
	3. We next prove that the sequences of $\lbrace \mathbf{X}_k \rbrace$ and $\lbrace \mathbf{Z}_k \rbrace$ are upper bounded.
	\begin{align}
		& \left\Vert \mathbf{W(Y-X)} \right\Vert_F^2 + \lambda (\Vert \mathbf{Z}_k \Vert_* - \alpha \Vert \mathbf{Z}_k \Vert_F ) \notag \\
		&= \mathcal{L}_{\rho_{k-1}} \left( \mathbf{X}_{k}, \mathbf{Z}_{k}, \mathbf{A}_{k-1} \right) - \left \langle \mathbf{A}_{k-1}, \mathbf{X}_k - \mathbf{Z}_k \right \rangle  - \frac{\rho_{k-1}}{2} \left \Vert \mathbf{X}_k - \mathbf{Z}_k \right \Vert_F^2 \notag \\
		&= \mathcal{L}_{\rho_{k-1}} \left( \mathbf{X}_{k}, \mathbf{Z}_{k}, \mathbf{A}_{k-1} \right) - \left \langle \mathbf{A}_{k-1}, (\mathbf{A}_k - \mathbf{A}_{k-1})/\rho_{k-1} \right \rangle  - \frac{\rho_{k-1}}{2} \left \Vert (\mathbf{A}_k - \mathbf{A}_{k-1})/\rho_{k-1} \right \Vert_F^2 \notag \\
		&= \mathcal{L}_{\rho_{k-1}} \left( \mathbf{X}_{k}, \mathbf{Z}_{k}, \mathbf{A}_{k-1} \right) + \frac{1}{2\rho_{k-1}} \left( \Vert \mathbf{A}_{k-1} \Vert_F^2 - \Vert \mathbf{A}_{k} \Vert_F^2 \right). \notag
	\end{align}
	Since both $\lbrace \mathcal{L}_{\rho_k} \left( \mathbf{X}_{k+1}, \mathbf{Z}_{k+1}, \mathbf{A}_k \right) \rbrace$ and $\lbrace \mathbf{A}_k \rbrace$ are upper bounded, we can deduce from the above that $\lbrace \mathbf{W(Y-X)} \rbrace$ and $\mathbf{Z}_k$ are upper bounded. Furthermore, since $\mathbf{X}_{k+1} = \mathbf{Z}_{k+1} + (\mathbf{A}_{k+1} - \mathbf{A}_{k})/\rho_k $, the sequence $\lbrace \mathbf{X}_k \rbrace$ is upper bounded. At last we have
	\begin{equation}
		\lim_{k \rightarrow \infty} \left\Vert \mathbf{X}_{k+1} - \mathbf{Z}_{k+1} \right \Vert_F = \lim_{k \rightarrow \infty} \rho_k^{-1} \left\Vert \mathbf{A}_{k+1} - \mathbf{A}_{k+1} \right\Vert_F = 0. \notag
	\end{equation}
	Therefore the stopping criterion (a) is proved.
	\par 

	4. In the forth step we prove the stopping criterion (b): $\lim_{k \rightarrow \infty}\Vert \mathbf{X}_{k+1} -\mathbf{X}_k \Vert_F = 0$.
	\begin{align}
		& \lim_{k \rightarrow \infty} \left \Vert \mathbf{X}_{k+1} -\mathbf{X}_k \right \Vert_F \notag \\
		& = \lim_{k \rightarrow \infty} \Vert ( \mathbf{W}^\top \mathbf{W} + \frac{\rho_k}{2} \mathbf{I} )^{-1} ( \mathbf{W}^\top \mathbf{WY} + \frac{\rho_k}{2}\mathbf{Z}_k - \frac{1}{2}\mathbf{A}_k ) - \rho_{k-1}^{-1} \left( \mathbf{A}_k - \mathbf{A}_{k-1} \right) - \mathbf{Z}_k \Vert_F \notag \\
		& = \lim_{k \rightarrow \infty} \Vert ( \mathbf{W}^\top \mathbf{W} + \frac{\rho_k}{2} \mathbf{I} )^{-1} (\mathbf{W}^\top \mathbf{WY} - \mathbf{W}^\top \mathbf{WZ}_k -\frac{1}{2}\mathbf{A}_k) - \rho_{k-1}^{-1}(\mathbf{A}_k - \mathbf{A}_{k-1}) \Vert_F \notag \\
		& \le \lim_{k \rightarrow \infty} \Vert ( \mathbf{W}^\top \mathbf{W} + \frac{\rho_k}{2} \mathbf{I} )^{-1} (\mathbf{W}^\top \mathbf{WY} + \mathbf{W}^\top \mathbf{WZ}_k -\frac{1}{2}\mathbf{A}_k) + \rho_{k-1}^{-1}(\mathbf{A}_k - \mathbf{A}_{k-1}) \Vert_F \ = 0. \notag
	\end{align}
	\par 
	5. Finally, we prove the stopping criterion (c). Given $\mathbf{Z}_{k+1} = \rho_k^{-1}(\mathbf{A}_k - \mathbf{A}_{k+1}) + \mathbf{X}_{k+1}$, $\mathbf{X}_k+\rho_{k-1}^{-1}\mathbf{A}_{k-1} = \mathbf{U}_{k-1} Diag([\sigma_1, \ldots, \sigma_M]^\top ) \mathbf{V}_{k-1}^\top$, and $\mathbf{Z}_k = \mathbf{U}_{k-1} Diag([\hat{\sigma}_1, \ldots, \hat{\sigma}_M]^\top ) \mathbf{V}_{k-1}^\top$, we have
	\begin{align}
		\lim_{k \rightarrow \infty} \left\Vert \mathbf{Z}_{k+1} -\mathbf{Z}_k \right\Vert_F &= \lim_{k \rightarrow \infty} \left\Vert \rho_k^{-1}(\mathbf{A}_k - \mathbf{A}_{k+1}) + \mathbf{X}_{k+1} - \mathbf{Z}_k \right\Vert_F \notag \\
		&= \lim_{k \rightarrow \infty} \Vert \mathbf{X}_k + \rho_{k-1}^{-1}\mathbf{A}_{k-1} - \mathbf{Z}_k + \mathbf{X}_{k+1} - \mathbf{X}_k \notag - \rho_{k-1}^{-1}\mathbf{A}_{k-1} + \rho_k^{-1}(\mathbf{A}_k - \mathbf{A}_{k+1}) \Vert_F \notag \\
		& \le \lim_{k \rightarrow \infty} \Vert Diag([\sigma_1, \ldots, \sigma_M]^\top ) - Diag([\hat{\sigma}_1, \ldots, \hat{\sigma}_M]^\top ) \Vert_F + \left\Vert \mathbf{X}_{k+1} - \mathbf{X}_k \right\Vert_F \notag \\
		& \quad \; + \left\Vert \rho_{k-1}^{-1}\mathbf{A}_{k-1} + \rho_{k}^{-1}(\mathbf{A}_{k+1} - \mathbf{A}_{k}) \right\Vert_F \notag \\
		& \le \lim_{k \rightarrow \infty} \textstyle \sqrt{\sum_{i=1}^{M} \left( \lambda/\rho_k \right)^2} + \left\Vert \mathbf{X}_{k+1} - \mathbf{X}_k \right\Vert_F + \left\Vert \rho_{k-1}^{-1}\mathbf{A}_{k-1} + \rho_{k}^{-1}\mathbf{A}_{k+1} - \rho_{k}^{-1}\mathbf{A}_{k} \right\Vert_F \ = 0 \notag
	\end{align}
	Up to now, the stopping criteria (a), (b) and (c) are all proved.  \qed
	\par 
	Theorem \ref{Th1} ensures that the difference between variables $\mathbf{X}$ and $\mathbf{Z}$ tends towards zero. And the changes of $\mathbf{X}$ and $\mathbf{Z}$ in consecutive iterations both tend towards zero. Since the variable sequences generated by algorithm \ref{admm} are bound to converge to their respective stationary points (except $\lbrace \rho_k \rbrace$), the convergence guarantee of algorithm \ref{admm} is established.
	\par 
	We discuss the computational complexity of the MC-NNFNM model in brief. In Algorithm \ref{admm}, updating $\mathbf{X}$ and $\mathbf{Z}$ cost $\mathcal{O}( p^4M )$ and $\mathcal{O}(p^4M + p^2M^2)$, respectively. The costs of updating $\mathbf{A}$ and $\rho$ are $\mathcal{O}(p^2M)$ and $\mathcal{O}(1)$, respectively. Therefore, the overall complexity of the proposed MC-NNFNM model is $\mathcal{O}((p^4M + p^2M^2)K_1)$. In Algorithm \ref{framework}, step 3 and 9 cost $\mathcal{O} (3p^2 S^2)$ and $\mathcal{O} (3p^2 m n)$, respectively. Note that step 5 and 6 are executed $K_2N$ times, and each execution of them cost $\mathcal{O} (S^2 log S)$ and $\mathcal{O}((p^4M + p^2M^2)K_1)$, respectively. Thus the dominant cost lies in step 6. And the overall complexity of Algorithm \ref{framework} is $\mathcal{O}((p^4M + p^2M^2)K_1K_2N)$.
	
	\section{Experimental Results}
	The performance of the proposed MC-NNFNM is evaluated on synthetic and real noise datasets. We compare MC-NNFNM with seven state-of-the-art methods, including the CBM3D \cite{BM3D}, MCWNNM \cite{MCWNNM}, WCWSNM \cite{MCWSNM}, denoising convolutional neural networks (DnCNN) \cite{DnCNN}, fast and flexible denoising network (FFDNet) \cite{FFDNet}, guided image denoisng (GID) \cite{GID} and Neat Image \cite{NI}. Specifically, CBM3D is one of the benchmark methods for color image denoising. The results obtained yield by it are used as baseline comparison. MCWNNM and MCWSNM are excellent low-rank minimization methods. DnCNN and FFDNet are repersentitive CNN-based methods. GID is an competitive guided dictionary learning methods for real-world image denoising. Neat Image (NI) is a commercial software with throughly optimized algorithms. All the experiments (except that of Neat Image) in this paper are implemented in MATLAB R2020a on a laptop (Windows 10, 2.1GHz CPU, 16GB RAM, Nvidia GeForce MX350 GPU). While the experiments of Neat Image are implemented in Adobe Photoshop CC 2019 on the same laptop.
	
	\begin{figure}[t] 
		\captionsetup{captionskip = 0pt}
		\centering
		\subfloat{
			\hspace{-2.1mm}
			\includegraphics[width=0.08\textwidth]{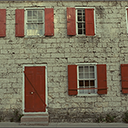}
			\hspace{-2.1mm}
			\includegraphics[width=0.08\textwidth]{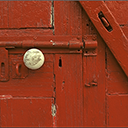}
			\hspace{-2.1mm}
			\includegraphics[width=0.08\textwidth]{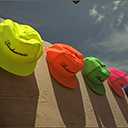}
			\hspace{-2.1mm}
			\includegraphics[width=0.08\textwidth]{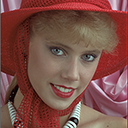}
			\hspace{-2.1mm}
			\includegraphics[width=0.08\textwidth]{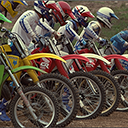}
			\hspace{-2.1mm}
			\includegraphics[width=0.08\textwidth]{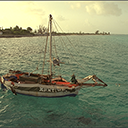}
			\hspace{-2.1mm}
			\includegraphics[width=0.08\textwidth]{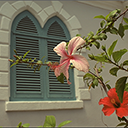}
			\hspace{-2.1mm}
			\includegraphics[width=0.08\textwidth]{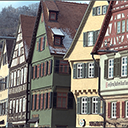}
			\hspace{-2.1mm}
			\includegraphics[width=0.08\textwidth]{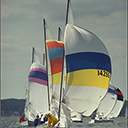}
			\hspace{-2.1mm}
			\includegraphics[width=0.08\textwidth]{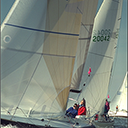}
			\hspace{-2.1mm}
			\includegraphics[width=0.08\textwidth]{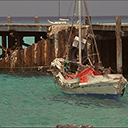}
			\hspace{-2.1mm}
			\includegraphics[width=0.08\textwidth]{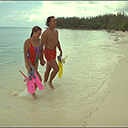}
			\hspace{-2.1mm}
		}%
		\vspace{0mm}
		\subfloat{
			\hspace{-2.1mm}
			\includegraphics[width=0.08\textwidth]{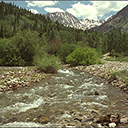}
			\hspace{-2.1mm}
			\includegraphics[width=0.08\textwidth]{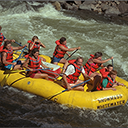}
			\hspace{-2.1mm}
			\includegraphics[width=0.08\textwidth]{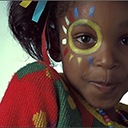}
			\hspace{-2.1mm}
			\includegraphics[width=0.08\textwidth]{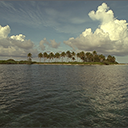}
			\hspace{-2.1mm}
			\includegraphics[width=0.08\textwidth]{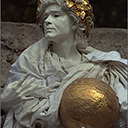}
			\hspace{-2.1mm}
			\includegraphics[width=0.08\textwidth]{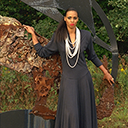}
			\hspace{-2.1mm}
			\includegraphics[width=0.08\textwidth]{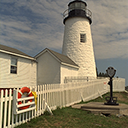}
			\hspace{-2.1mm}
			\includegraphics[width=0.08\textwidth]{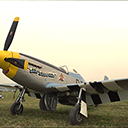}
			\hspace{-2.1mm}
			\includegraphics[width=0.08\textwidth]{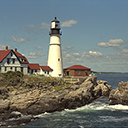}
			\hspace{-2.1mm}
			\includegraphics[width=0.08\textwidth]{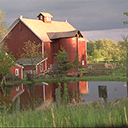}
			\hspace{-2.1mm}
			\includegraphics[width=0.08\textwidth]{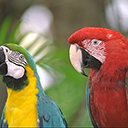}
			\hspace{-2.1mm}
			\includegraphics[width=0.08\textwidth]{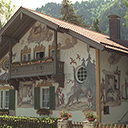}
			\hspace{-2.1mm}
		}%
		\caption{\centering Twenty-four test images in Kodak PhotoCD dataset (enumerated from left-to-right and
top-to-bottom).}
		\label{thumb_kodak}
	\end{figure}
	\begin{figure}[tb]
		\centering
		\includegraphics[width=8cm]{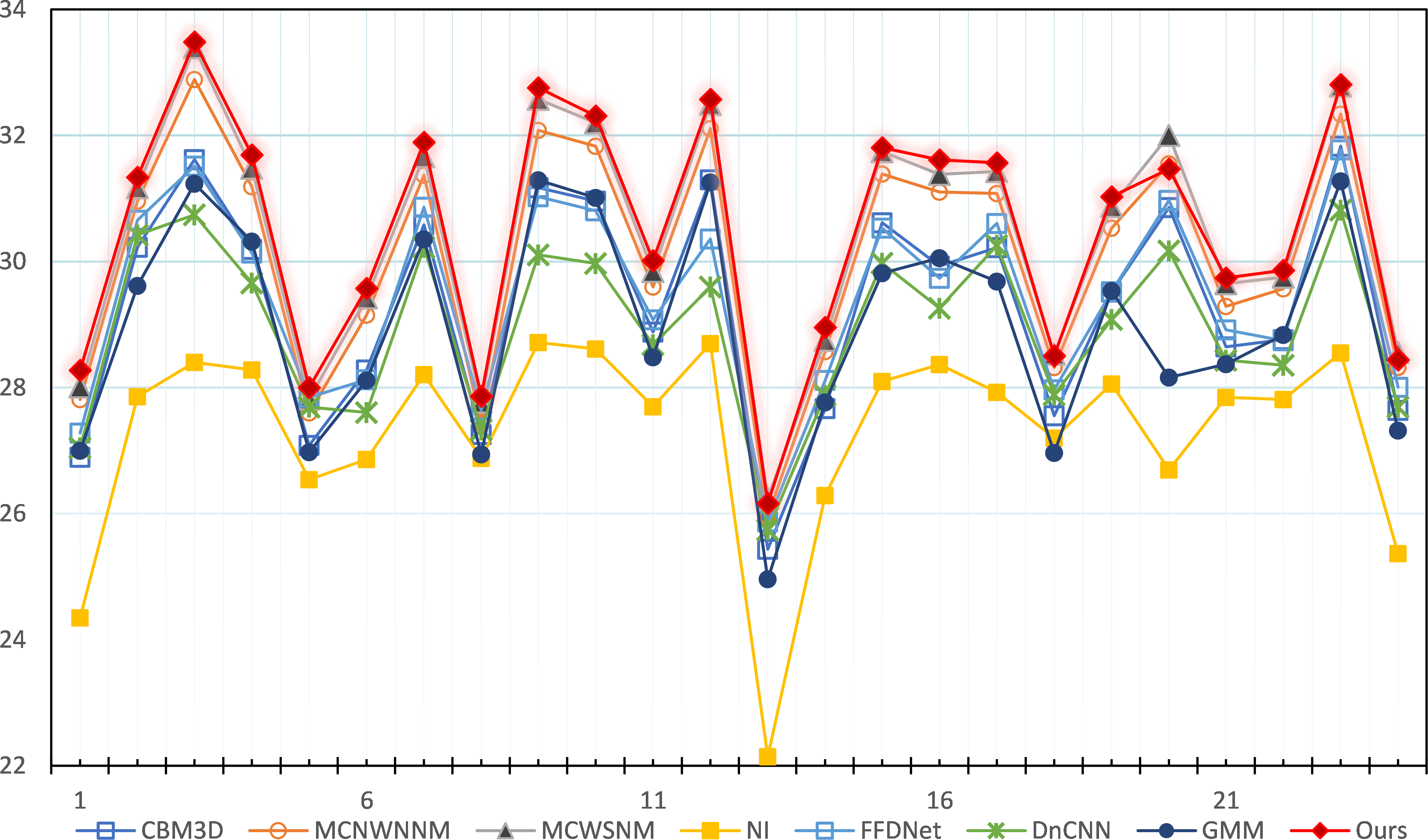}
		\caption{\centering Line chart of the PSNR(dB) results for all competing methods on Kodak PhotoCD dataset.}
		\label{line_syn}
	\end{figure}
	\subsection{Experimental Settings}
	In synthetic noise experiments, the noise levels in each channel are assumed to be known. Zero-mean AWGN with variances $(\sigma_r^2, \sigma_g^2, \sigma_b^2)$ are added to each channel to generate noisy observations. In real noise experiments, the noise levels of noisy observations are assumed to be AWGN and can be estimated by the noise estimation algorithm in \cite{noisEstimate}. The CBM3D, DnCNN and FFDNet receive a single inputted noise level. We set it to be the mean of the noise variances in three channels,i.e., $ \sigma^2 = (\sigma_r^2 + \sigma_g^2 + \sigma_b^2)/3 $. For GID, we tune its parameter $c1$ to meet its best performance for synthetic experiments. We also tune the parameters $(\lambda, \rho)$ of MCWSNM for real noise experiments. For other competing methods, we keep their default parameters mentioned in original papers.
	
    \begin{table*}[pt] 
		\caption{\newline PSNR(dB) results for all competing methods on Kodak PhotoCD dataset.}
		\begin{tabular}{ccccccccp{1.87cm}<{\centering}}
			\toprule
			\multicolumn{9}{c}{$\sigma_r = 30,\quad \sigma_g = 10,\quad \sigma_b = 50$} \\
			\midrule
			Image\# & \textbf{CBM3D} & \textbf{MCWNNM} & \textbf{MCWSNM} & \textbf{NI} & \textbf{FFDNet} & \textbf{DnCNN} & \textbf{GID} & \textbf{Ours} \\
			\midrule
			1  &26.90 &27.81 &28.01 &24.35 &27.28 &27.04 &27.00 &\textbf{28.34}  \\
			2  &30.23 &30.96 &31.17 &27.86 &30.66 &30.42 &29.61 &\textbf{31.40}  \\
			3  &31.62 &32.89 &33.39 &28.40 &31.51 &30.74 &31.23 &\textbf{33.59}  \\
			4  &30.19 &31.19 &31.48 &28.28 &30.13 &29.66 &30.32 &\textbf{31.76}  \\
			5  &27.08 &27.60 &27.82 &26.54 &27.85 &27.69 &26.98 &\textbf{28.09}  \\
			6  &28.28 &29.15 &29.43 &26.86 &28.12 &27.61 &28.11 &\textbf{29.66}  \\
			7  &30.58 &31.37 &31.66 &28.21 &30.86 &30.23 &30.35 &\textbf{32.02}  \\
			8  &27.25 &27.44 &27.77 &26.88 &27.65 &27.33 &26.94 &\textbf{27.89}  \\
			9  &31.17 &32.08 &32.57 &28.71 &31.03 &30.11 &31.29 &\textbf{32.84}  \\
			10 &30.96 &31.83 &32.20 &28.61 &30.81 &29.97 &31.01 &\textbf{32.47}  \\
			11 &28.88 &29.60 &29.84 &27.70 &29.08 &28.68 &28.48 &\textbf{30.09}  \\
			12 &31.30 &32.11 &32.49 &28.70 &30.36 &29.60 &31.26 &\textbf{32.63}  \\
			13 &25.44 &25.96 &26.25 &22.14 &25.87 &25.74 &24.96 &\textbf{26.25}  \\
			14 &27.67 &28.57 &28.75 &26.29 &28.12 &27.88 &27.76 &\textbf{29.00}  \\
			15 &30.62 &31.39 &31.75 &28.10 &30.53 &29.97 &29.82 &\textbf{31.88}  \\
			16 &29.93 &31.10 &31.38 &28.36 &29.74 &29.26 &30.06 &\textbf{31.67}  \\
			17 &30.22 &31.08 &31.43 &27.92 &30.60 &30.25 &29.68 &\textbf{31.62}  \\
			18 &27.55 &28.32 &28.53 &27.20 &27.96 &27.89 &26.96 &\textbf{28.57}  \\
			19 &29.52 &30.53 &30.87 &28.06 &29.52 &29.08 &29.53 &\textbf{31.06}  \\
			20 &30.85 &31.55 &\textbf{32.00} &26.69 &30.97 &30.17 &28.16 &31.57  \\
			21 &28.65 &29.29 &29.65 &27.84 &28.91 &28.44 &28.37 &\textbf{29.74}  \\
			22 &28.76 &29.57 &29.75 &27.81 &28.75 &28.36 &28.84 &\textbf{29.88}  \\
			23 &31.83 &32.34 &32.78 &28.55 &31.77 &30.82 &31.27 &\textbf{32.91}  \\
			24 &27.64 &28.32 &\textbf{28.56} &25.36 &28.00 &27.69 &27.32 &28.48  \\
			\midrule
			\textbf{Avg} &29.30 &30.09 &30.40 &27.31 &29.42&28.94&28.97 &\textbf{30.56} \\
			\bottomrule
		\end{tabular}
		\label{synResults}
	\end{table*}
	\begin{figure}[pt] 
		\captionsetup[subfigure]{captionskip=0pt, farskip=0pt}
		\centering
		\subfloat[Ground Truth]{
			\hspace{-3mm}
			\includegraphics[width=0.192\textwidth]{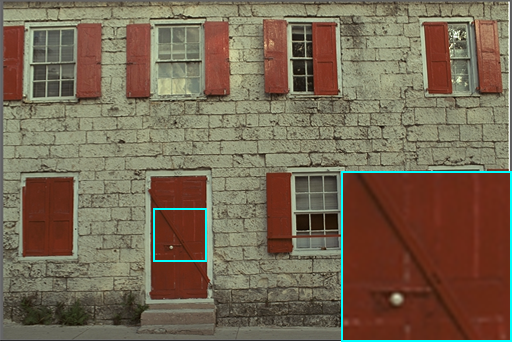}
			\hspace{-3mm}
		}
		\subfloat[Noisy]{
			\includegraphics[width=0.192\textwidth]{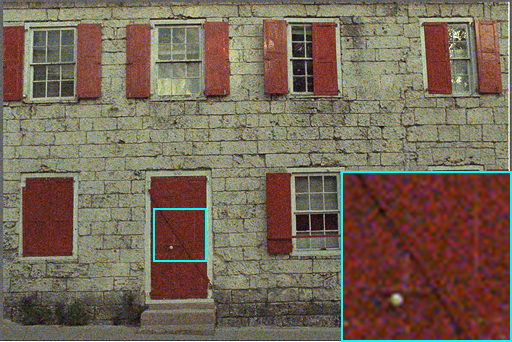}
			\hspace{-3mm}
		}%
		\subfloat[\scriptsize{CBM3D: 26.90dB}]{
			\includegraphics[width=0.192\textwidth]{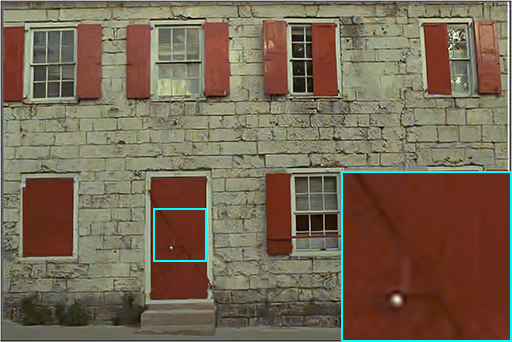}
			\hspace{-3mm}
		}%
		\subfloat[\scriptsize{MCWNNM: 27.81dB}]{
			\includegraphics[width=0.192\textwidth]{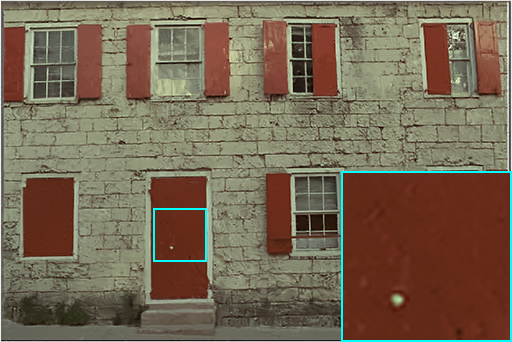}
			\hspace{-3mm}
		}%
		\subfloat[\scriptsize{MCWSNM: 28.01dB}]{
			\includegraphics[width=0.192\textwidth]{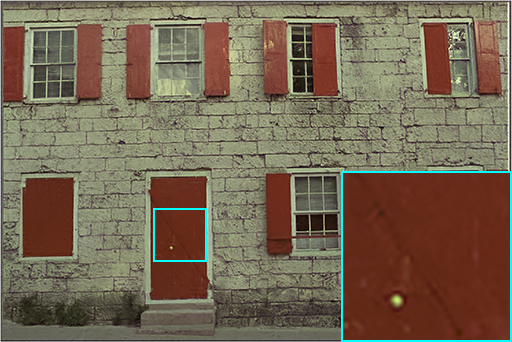}
			\hspace{-3mm}
		}%
		
		\subfloat[\scriptsize{NI: 24.35dB}]{
			\hspace{-3mm}
			\includegraphics[width=0.192\textwidth]{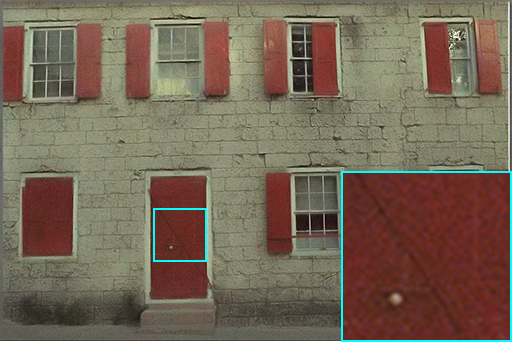}
			\hspace{-3mm}
		}%
		\subfloat[FFDNet: 27.28dB]{
			\includegraphics[width=0.192\textwidth]{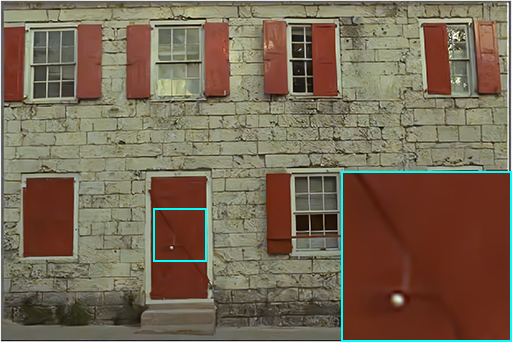}
			\hspace{-3mm}
		}%
		\subfloat[\scriptsize{DnCNN: 27.04dB}]{
			\includegraphics[width=0.192\textwidth]{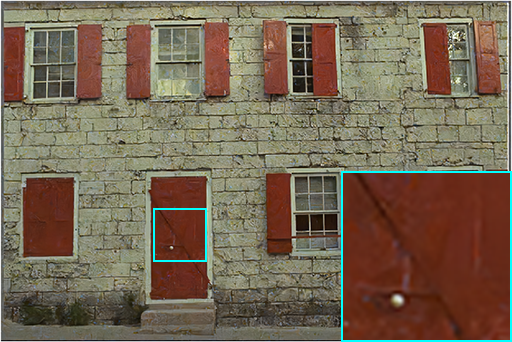}
			\hspace{-3mm}
		}%
		\subfloat[\scriptsize{GID: 27.00dB}]{
			\includegraphics[width=0.192\textwidth]{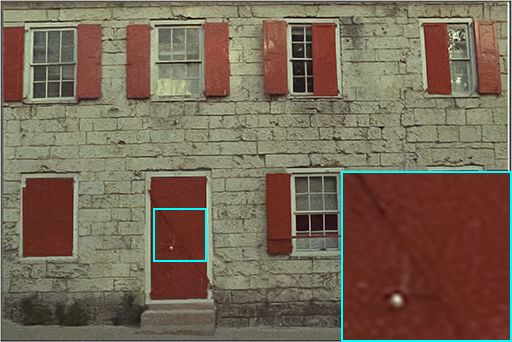}
			\hspace{-3mm}
		}%
		\subfloat[\scriptsize{\textbf{Ours: 28.34dB}}]{
			\includegraphics[width=0.192\textwidth]{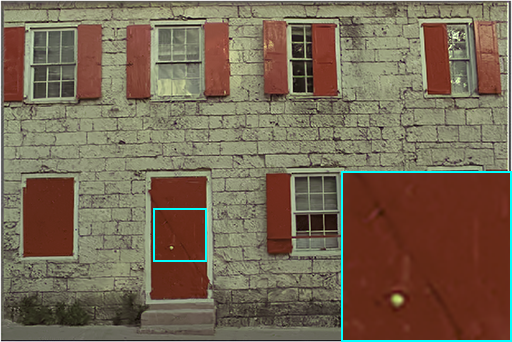}
			\hspace{-3mm}
		}%
		\caption{\centering Denoised results on image ``kodim01'' with PSNR(dB) results.}
		\label{visual_syn_1}
	\end{figure}
	\begin{figure}[p] 
		\captionsetup[subfigure]{captionskip=0pt, farskip=0pt}
		\centering
		\subfloat[Ground Truth]{
			\hspace{-3mm}
			\includegraphics[width=0.192\textwidth]{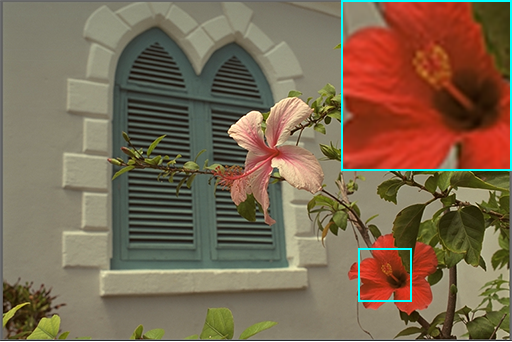}
			\hspace{-3mm}
		}
		\subfloat[Noisy]{
			\includegraphics[width=0.192\textwidth]{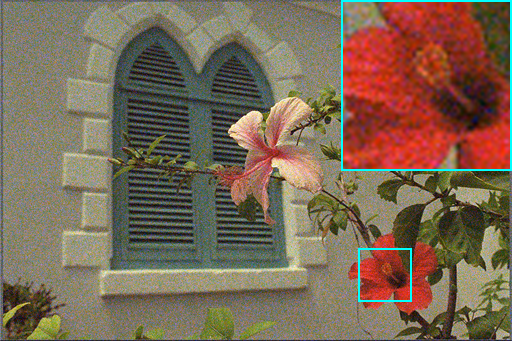}
			\hspace{-3mm}
		}%
		\subfloat[\scriptsize{CBM3D: 30.58dB}]{
			\includegraphics[width=0.192\textwidth]{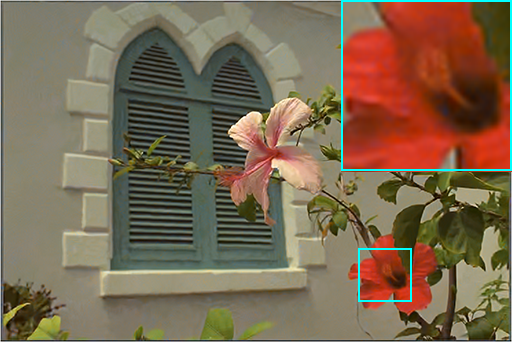}
			\hspace{-3mm}
		}%
		\subfloat[\scriptsize{MCWNNM: 31.37dB}]{
			\includegraphics[width=0.192\textwidth]{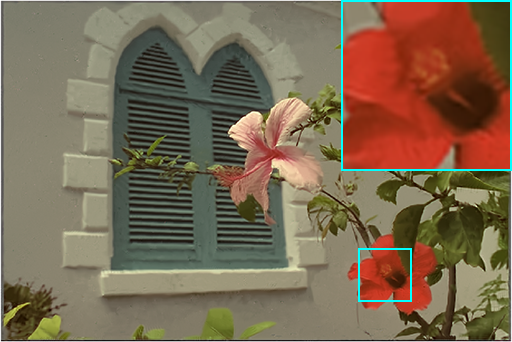}
			\hspace{-3mm}
		}%
		\subfloat[\scriptsize{MCWSNM: 31.66dB}]{
			\includegraphics[width=0.192\textwidth]{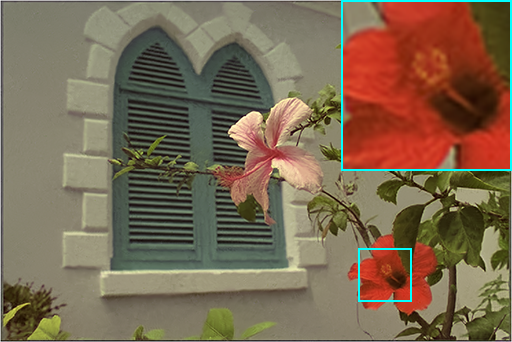}
			\hspace{-3mm}
		}%
		
		\subfloat[\scriptsize{NI: 28.21dB}]{
			\hspace{-3mm}
			\includegraphics[width=0.192\textwidth]{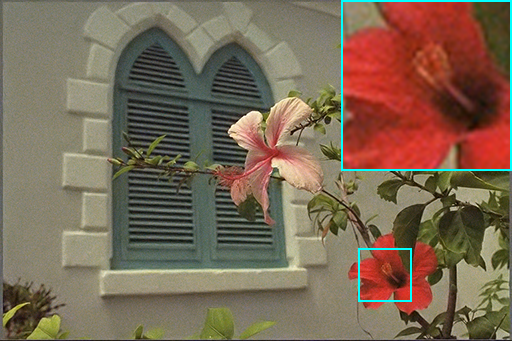}
			\hspace{-3mm}
		}%
		\subfloat[FFDNet: 30.86dB]{
			\includegraphics[width=0.192\textwidth]{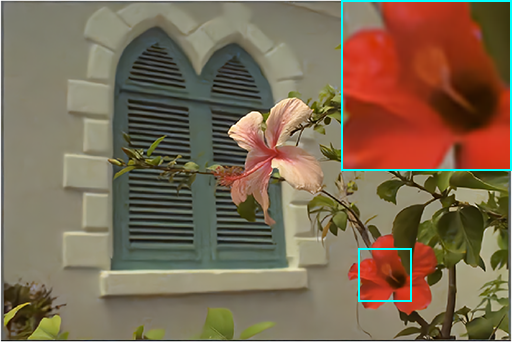}
			\hspace{-3mm}
		}%
		\subfloat[\scriptsize{DnCNN: 30.23dB}]{
			\includegraphics[width=0.192\textwidth]{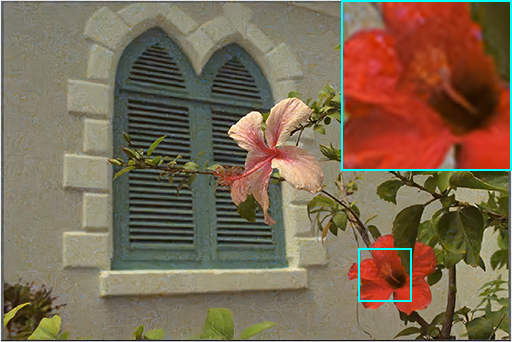}
			\hspace{-3mm}
		}%
		\subfloat[\scriptsize{GID: 30.35dB}]{
			\includegraphics[width=0.192\textwidth]{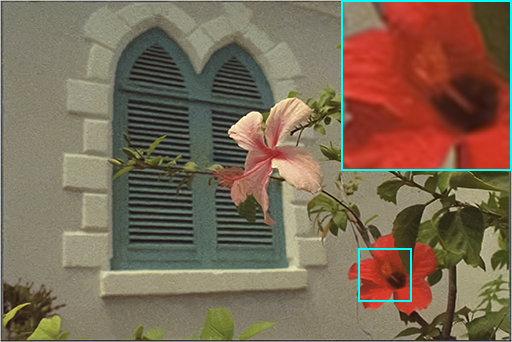}
			\hspace{-3mm}
		}%
		\subfloat[\scriptsize{\textbf{Ours: 32.02dB}}]{
			\includegraphics[width=0.192\textwidth]{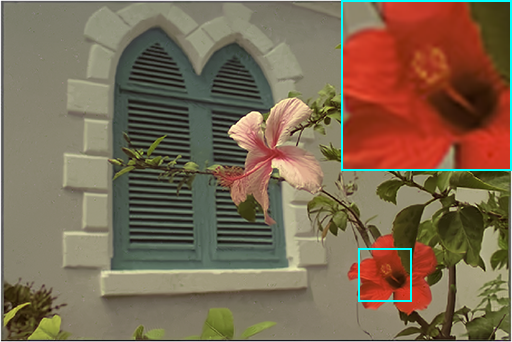}
			\hspace{-3mm}
		}%
		\caption{\centering Denoised results on image ``kodim07'' with PSNR(dB) results. }
		\label{visual_syn_7}
	\end{figure}
	\begin{figure}[t] 
		\captionsetup[subfigure]{captionskip=0pt, farskip=0pt}
		\centering
		\subfloat[Ground Truth]{
			\hspace{-3mm}
			\includegraphics[width=0.192\textwidth]{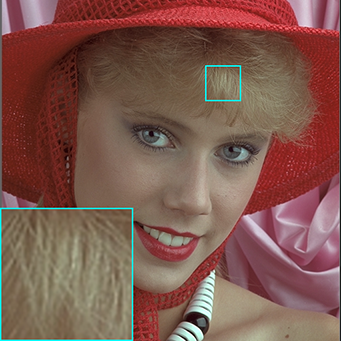}
			\hspace{-3mm}
		}
		\subfloat[Noisy]{
			\includegraphics[width=0.192\textwidth]{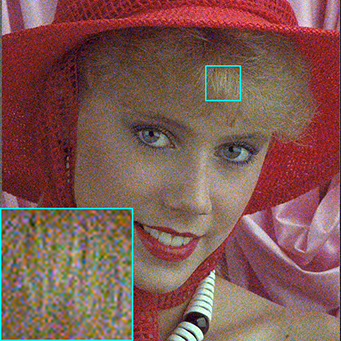}
			\hspace{-3mm}
		}%
		\subfloat[\scriptsize{CBM3D: 30.19dB}]{
			\includegraphics[width=0.192\textwidth]{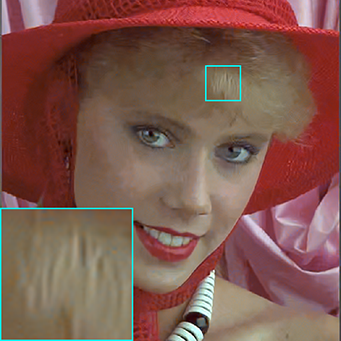}
			\hspace{-3mm}
		}%
		\subfloat[\scriptsize{MCWNNM: 31.19dB}]{
			\includegraphics[width=0.192\textwidth]{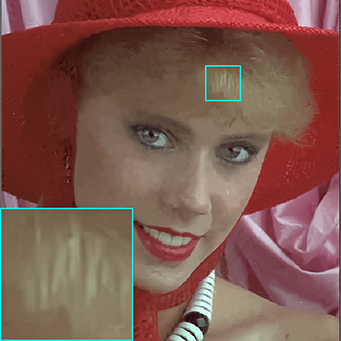}
			\hspace{-3mm}
		}%
		\subfloat[\scriptsize{MCWSNM: 31.48dB}]{
			\includegraphics[width=0.192\textwidth]{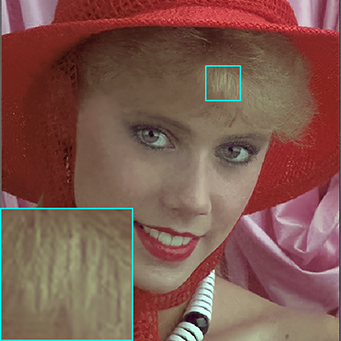}
			\hspace{-3mm}
		}%
		
		\subfloat[\scriptsize{NI: 28.28dB}]{
			\hspace{-3mm}
			\includegraphics[width=0.192\textwidth]{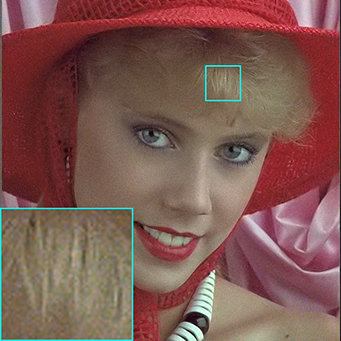}
			\hspace{-3mm}
		}%
		\subfloat[FFDNet: 30.13dB]{
			\includegraphics[width=0.192\textwidth]{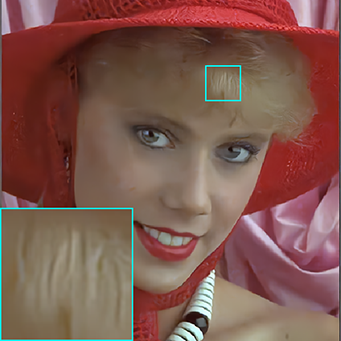}
			\hspace{-3mm}
		}%
		\subfloat[\scriptsize{DnCNN: 29.66dB}]{
			\includegraphics[width=0.192\textwidth]{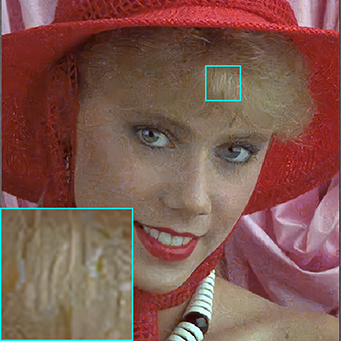}
			\hspace{-3mm}
		}%
		\subfloat[\scriptsize{GID: 30.32dB}]{
			\includegraphics[width=0.192\textwidth]{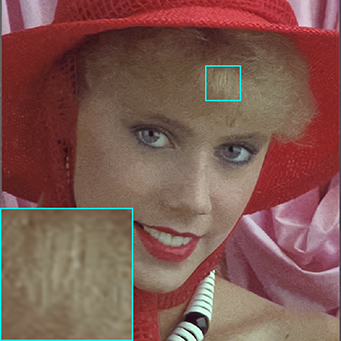}
			\hspace{-3mm}
		}%
		\subfloat[\scriptsize{\textbf{Ours: 31.76dB}}]{
			\includegraphics[width=0.192\textwidth]{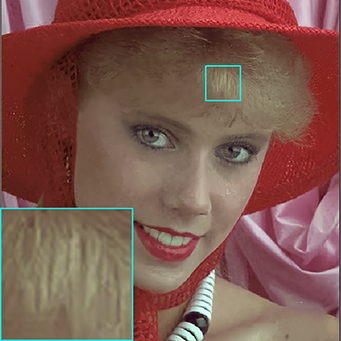}
			\hspace{-3mm}
		}%
		\caption{\centering Denoised results on image ``kodim04'' with PSNR(dB) results. }
		\label{visual_syn_4}
	\end{figure}

	\subsection{Experiments on the Synthetic Noise Dataset}
	We compare MC-NNFNM with other competing methods on the Kodak PhotoCD dataset, which involves twenty-four high quality color images. Its thumbnails are shown in \figref{thumb_kodak}. All competing methods are tested under the noise levels $(\sigma_r, \sigma_g, \sigma_b)=(30, 10, 50)$. For MC-NNFNM, we set the search window size as 20, each patch size $p=6$, the iteration number $K_2=5$, the trade-off parameter $\lambda=0.86$, the iteration number for ADMM $K_1 = 10$, the penalize parameter $\rho_0 = 0.86$, $\mu_0 = 1.001$, and $\alpha = 1.9$. 
	\par 
	The PSNR results for all competing methods are shown in \figref{line_syn} and Table \ref{synResults}. The highest results for each image are highlighted in bold. Table \ref{synResults} shows that the proposed MC-NNFNM achieves the highest PSNR in 22 out of 24 images. On average, MC-NNFNM achieves 0.47dB and 0.16dB improvements over the MCWNNM and MCWSNM, respectively. Thus the proposed MC-NNFNM outperforms all the other methods at this noise level. In terms of visual comparision, as shown in \figref{visual_syn_1} $\sim$ \figref{visual_syn_4}, our method is capable of reconstructing more image details from the noisy observation. In the demarcated window of \figref{visual_syn_7}, MC-NNFNM well reconstructs the tiny pistils while reducing the noise. And in the demarcated window of \figref{visual_syn_4}, MC-NNFNM removes the noise completely without damaging the human hair details. In comparison, the CBM3D, MCWNNM and FFDNet over-smooth the image, while the MCWSNM, NI and GID did not remove the noise completely. And  To sum up, MC-NNFNM shows strong denoising capacity, producing pleasant visual quality while holding higher PSNR indices.

\subsection{Experiments on the Real Noise Dataset}
\begin{figure*}[t] 
		\centering
		\captionsetup[subfigure]{captionskip=0pt, farskip=-2pt}
			\subfloat{
			    \hspace{-2.1mm}
				\includegraphics[width=0.064\textwidth]{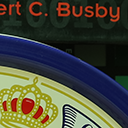}
				\hspace{-2.1mm}
				\includegraphics[width=0.064\textwidth]{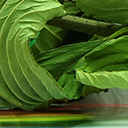}
				\hspace{-2.1mm}
				\includegraphics[width=0.064\textwidth]{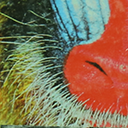}
				\hspace{-2.1mm}
				\includegraphics[width=0.064\textwidth]{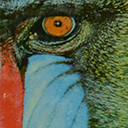}
				\hspace{-2.1mm}
				\includegraphics[width=0.064\textwidth]{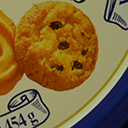}
				\hspace{-2.1mm}
				\includegraphics[width=0.064\textwidth]{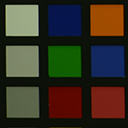}
				\hspace{-2.1mm}
				\includegraphics[width=0.064\textwidth]{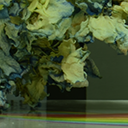}
				\hspace{-2.1mm}
				\includegraphics[width=0.064\textwidth]{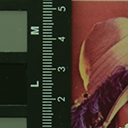}
				\hspace{-2.1mm}
				\includegraphics[width=0.064\textwidth]{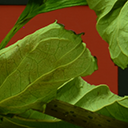}
				\hspace{-2.1mm}
				\includegraphics[width=0.064\textwidth]{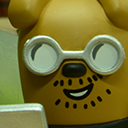}
				\hspace{-2.1mm}
				\includegraphics[width=0.064\textwidth]{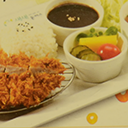}
				\hspace{-2.1mm}
				\includegraphics[width=0.064\textwidth]{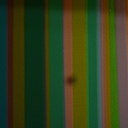}
				\hspace{-2.1mm}
				\includegraphics[width=0.064\textwidth]{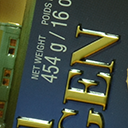}
				\hspace{-2.1mm}
				\includegraphics[width=0.064\textwidth]{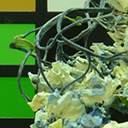}
				\hspace{-2.1mm}
				\includegraphics[width=0.064\textwidth]{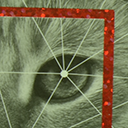}
			}
		\caption{\centering Fifteen test images in CC dataset (\#1 to \#15, enumerated from left-to-right).}
		\label{CC15}
	\end{figure*}

	We compare MC-NNFNM with other competing methods on the CC dataset \cite{CC15}. CC, whose thumbnails are shown in \figref{CC15} involves 15 cropped real-world noisy images and their corresponding clean version. The clean images are generated from 500 shots of the same scene using the same camera and camera settings. Hence they can be roughly regarded as the ground truth. With them, the PSNR can be computed and the quantitative comparision among competing methods can be implemented. 
    \par
\begin{figure}[p] 
		\captionsetup[subfigure]{captionskip=0pt, farskip=-2pt}
		\centering
		\subfloat[\scriptsize{Ground Truth}]{
			\hspace{-3mm}
			\includegraphics[width=0.183\textwidth]{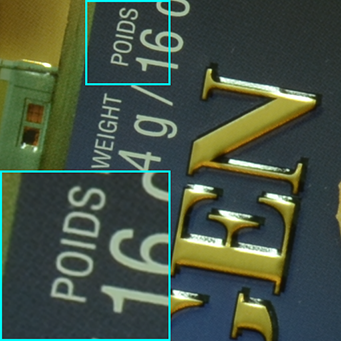}
			\hspace{-3mm}
		}%
		\subfloat[\scriptsize{Noisy}]{
			\includegraphics[width=0.183\textwidth]{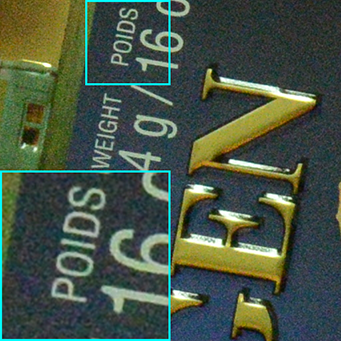}
			\hspace{-3mm}
		}%
		\subfloat[\scriptsize{CBM3D: 29.86dB}]{
			\includegraphics[width=0.183\textwidth]{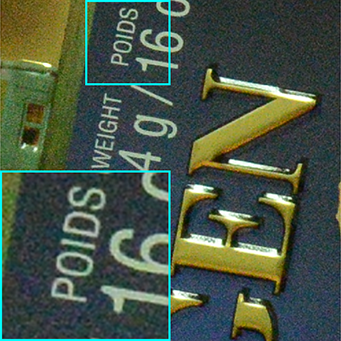}
			\hspace{-3mm}
		}%
		\subfloat[\scriptsize{MCWNNM:34.85dB}]{
			\includegraphics[width=0.183\textwidth]{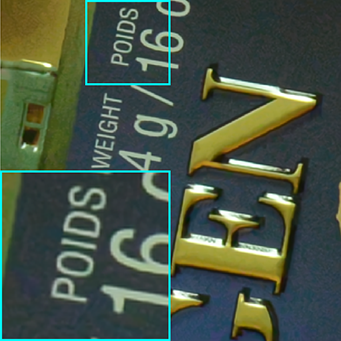}
			\hspace{-3mm}
		}%
		\subfloat[\scriptsize{MCWSNM:34.81dB}]{
			\includegraphics[width=0.183\textwidth]{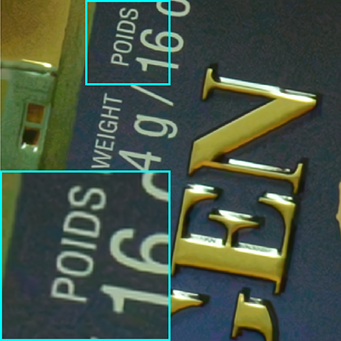}
			\hspace{-3mm}
		}%
		
		\subfloat[\scriptsize{NI: 33.61dB}]{
			\hspace{-3mm}
			\includegraphics[width=0.183\textwidth]{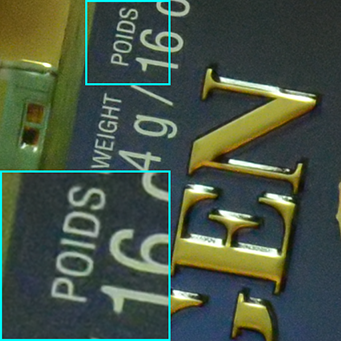}
			\hspace{-3mm}
		}%
		\subfloat[\scriptsize{FFDNet:29.79dB}]{
			\includegraphics[width=0.183\textwidth]{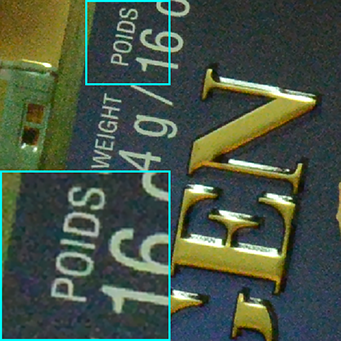}
			\hspace{-3mm}
		}%
		\subfloat[\scriptsize{DnCNN:30.02dB}]{
			\includegraphics[width=0.183\textwidth]{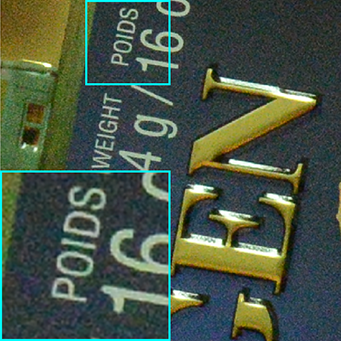}
			\hspace{-3mm}
		}%
		\subfloat[\scriptsize{GID: 33.01dB}]{
			\includegraphics[width=0.183\textwidth]{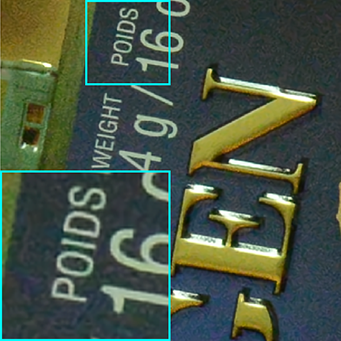}
			\hspace{-3mm}
		}%
		\subfloat[\scriptsize{\textbf{Ours: 34.89dB}}]{
			\includegraphics[width=0.183\textwidth]{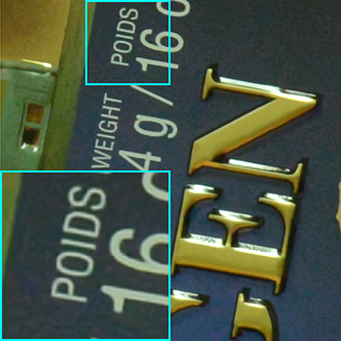}
			\hspace{-3mm}
		}%
		\caption{\centering Denoised results on image ``\#13'' with PSNR(dB) results. }
		\label{visual_real_13}
	\end{figure}

\begin{figure}[p] 
		\captionsetup[subfigure]{captionskip=0pt, farskip=-2pt}
		\centering
		\subfloat[\scriptsize{Ground Truth}]{
			\hspace{-3mm}
			\includegraphics[width=0.183\textwidth]{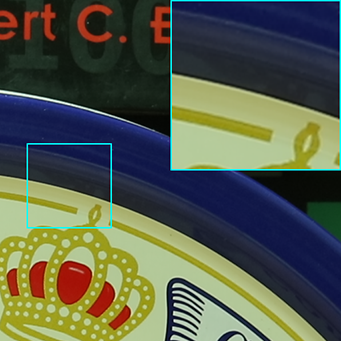}
			\hspace{-3mm}
		}%
		\subfloat[\scriptsize{Noisy}]{
			\includegraphics[width=0.183\textwidth]{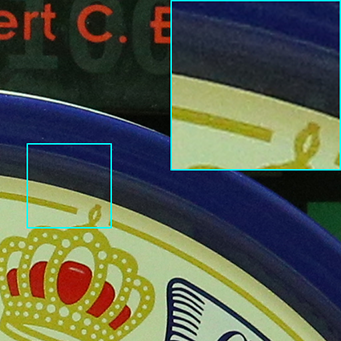}
			\hspace{-3mm}
		}%
		\subfloat[\scriptsize{CBM3D: 37.50dB}]{
			\includegraphics[width=0.183\textwidth]{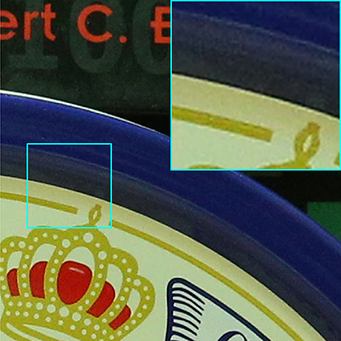}
			\hspace{-3mm}
		}%
		\subfloat[\scriptsize{MCWNNM: 41.22dB}]{
			\includegraphics[width=0.183\textwidth]{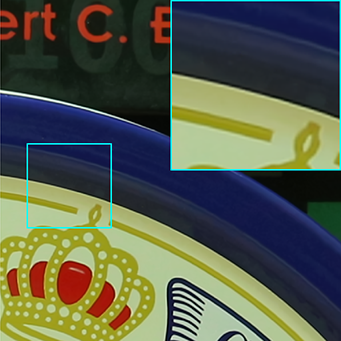}
			\hspace{-3mm}
		}%
		\subfloat[\scriptsize{MCWSNM: 40.80dB}]{
			\includegraphics[width=0.183\textwidth]{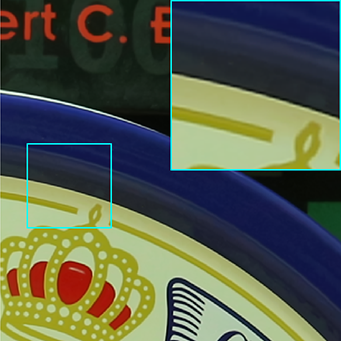}
			\hspace{-3mm}
		}%
		
		\subfloat[\scriptsize{NI: 37.72dB}]{
			\hspace{-3mm}
			\includegraphics[width=0.183\textwidth]{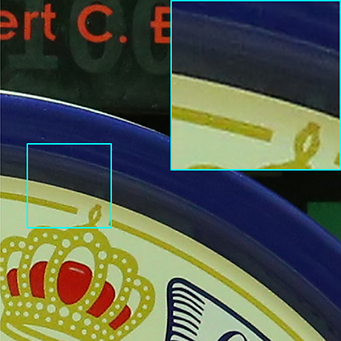}
			\hspace{-3mm}
		}%
		\subfloat[\scriptsize{FFDNet: 37.63dB}]{
			\includegraphics[width=0.183\textwidth]{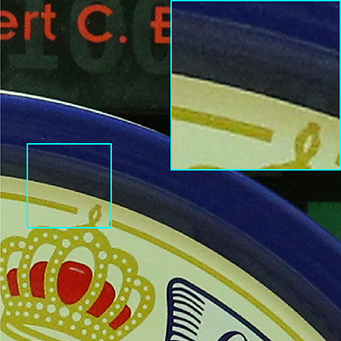}
			\hspace{-3mm}
		}%
		\subfloat[\scriptsize{DnCNN: 37.62dB}]{
			\includegraphics[width=0.183\textwidth]{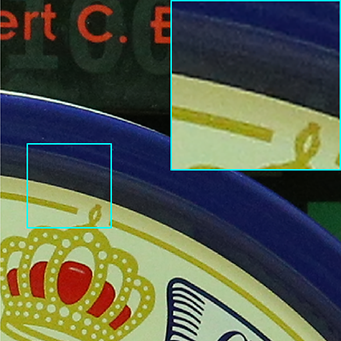}
			\hspace{-3mm}
		}%
		\subfloat[\scriptsize{GID: 40.82dB}]{
			\includegraphics[width=0.183\textwidth]{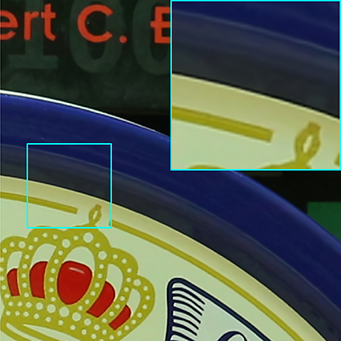}
			\hspace{-3mm}
		}%
		\subfloat[\scriptsize{\textbf{Ours: 41.35dB}}]{
			\includegraphics[width=0.183\textwidth]{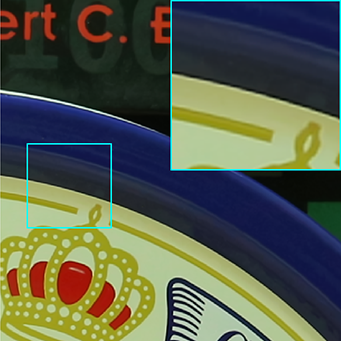}
			\hspace{-3mm}
		}%
		\caption{\centering Denoised results on image ``\#1'' with PSNR(dB) results. }
		\label{visual_real_1}
	\end{figure}

 
	For MC-NNFNM, we tune $\lambda = 4.86$, $\rho_0 = 4.55$ and $\alpha = 1.05$. Other parameters are kept the same as the synthetic noise experiments. The PSNR results for competing methods are listed in Table \ref{realResults}. The highest results for each image are highlighted in bold. Our method achieves highest PSNR in 10 out of 15 images. For visual comparison, as shown in \figref{visual_real_13} and \figref{visual_real_1}, our method produces promising visual quality over other competing methods. In the demarcated window of \figref{visual_real_13}, MC-NNFNM not only reduces the noise completely but also preserves the major textures and image details. In comparison, the CBM3D, FFDNet, DnCNN and GID remain the noise, while the NI and MCWNNM over-smooth the image. In summary, MC-NNFNM presents competitive denoising capability and produces satisfactory denoising results in visualization.
	\begin{table}[p] 
		\caption{\newline PSNR(dB) results of all competing methods on real noise CC dataset}
		\begin{tabular}{rccccccccp{1.3cm}<{\centering}}
			\toprule
			\makecell{Camera\\Settings} & {\#} & {CBM3D} & {MCWNNM} & {MCWSNM} & {NI} & {FFDNet} & {DnCNN} & {GID} & \textbf{Ours} \\
			\midrule
			\multirow{3}{*}{\makecell[r]{Ganon 5D,\\ ISO = 3200}} & 1 &37.50 &41.22 &40.80 &37.72 &37.63 &37.62 &40.82 &\textbf{41.35} \\
			&2 &34.33 &{37.25} &\textbf{37.34} &35.26 &34.51 &34.48 &37.19 &{37.16} \\
			&3 &34.37 &36.48 &{36.99} &34.89 &34.60 &34.65 &\textbf{36.92} &{36.25} \\
			\midrule
			\multirow{3}{*}{\makecell[r]{Nikon D600,\\ ISO = 3200}} &4 &33.44 &\textbf{35.54} &35.28 &34.70 &33.50 &33.48 &35.32 &{35.52} \\
			&5 &34.01 &37.03 &36.66 &34.32 &34.09 &34.16 &36.62 &\textbf{37.04} \\
			&6 &35.26 &39.56 &{39.53} &38.57 &35.38 &35.43 &38.68 &\textbf{39.57} \\
			\midrule
			\multirow{3}{*}{\makecell[r]{Nikon D800,\\ ISO = 1600}} &7 &35.78 &\textbf{39.26} &39.07 &38.18 &35.94 &35.93 &38.88 &{39.21} \\
			&8 &36.10 &41.45 &41.15 &38.85 &36.28 &36.28 &40.66 &\textbf{41.45} \\
			&9 &35.30 &\textbf{39.54} &39.39 &38.44 &35.30 &35.36 &39.20 &39.50 \\
			\midrule
			\multirow{3}{*}{\makecell[r]{Nikon D800,\\ ISO = 3200}} &10 &33.53 &38.94 &{38.89} &37.43 &33.61 &33.68 &37.92 &\textbf{38.99} \\
			&11 &33.06 &37.40 &{37.23} &35.72 &33.16 &33.16 &36.62 &\textbf{37.40} \\
			&12 &33.24 &39.42 &{39.51} &38.58 &33.34 &33.36 &37.64 &\textbf{39.45} \\
			\midrule
			\multirow{3}{*}{\makecell[r]{Nikon D800,\\ ISO = 6400}} &13 &29.86 &34.85 &34.47 &33.61 &29.79 &30.02 &33.01 &\textbf{34.89} \\
			&14 &30.20 &33.97 &33.56 &32.57 &30.33 &30.32 &32.93 &\textbf{33.98} \\
			&15 &30.02 &33.96 &{33.78} &32.86 &30.11 &30.13 &32.96 &\textbf{34.06} \\
			\midrule
			\textbf{Avg}& &33.73 &\textbf{37.72} &37.54 &36.11 &33.85 &33.87  &37.03 &37.71 \\
			\bottomrule
		\end{tabular}
		\label{realResults}
	\end{table}
	
	\begin{figure}[th] 
		\centering 
		\includegraphics[width = 8cm]{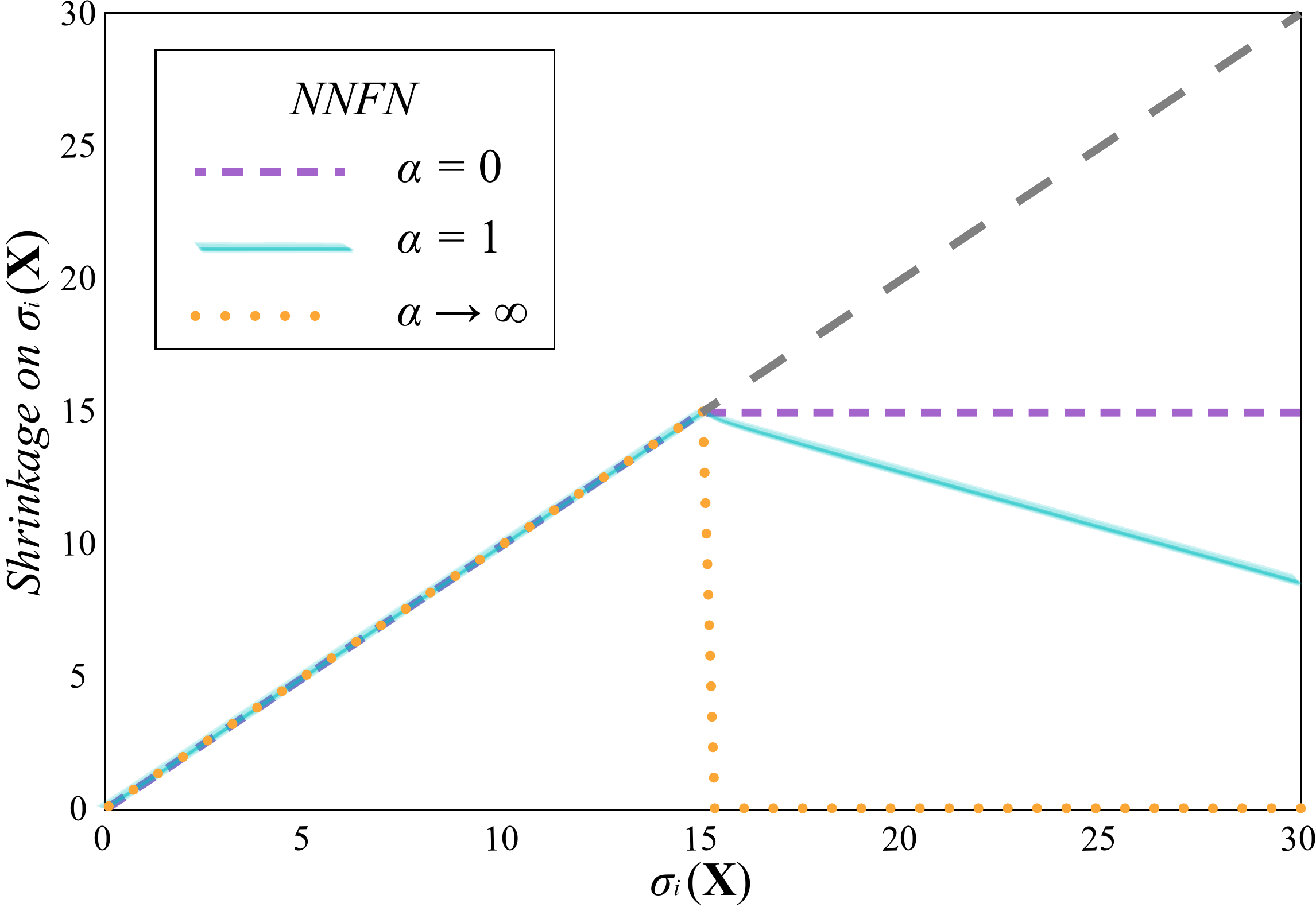}
		\caption{\centering Shrinkage performed by NNFN regularizer with different $\alpha$.}
		\label{NNFN_alpha}
	\end{figure}
	
	\begin{figure*}[hbt] 
		\centering
		\captionsetup[subfigure]{captionskip=0pt, farskip=0pt}
		\begin{minipage}{0.6\textwidth}
			\subfloat[kodim13: 26.00dB]{
				\includegraphics[width=0.32\textwidth]{img/kodim/kodim13.png}
				\hspace{-3mm}
			}
			\subfloat[\textbf{kodim08}: 27.77dB]{
				\includegraphics[width=0.32\textwidth]{img/kodim/kodim08.png}
				\hspace{-3mm}
			}
			\subfloat[kodim05: 27.91dB]{
				\includegraphics[width=0.32\textwidth]{img/kodim/kodim05.png}
				\hspace{-3mm}
			}\\
			\subfloat[\textbf{kodim03}: 33.50dB]{
				\includegraphics[width=0.32\textwidth]{img/kodim/kodim03.png}
				\hspace{-3mm}
			}
			\subfloat[ kodim23: 32.83dB]{
				\includegraphics[width=0.32\textwidth]{img/kodim/kodim23.png}
				\hspace{-3mm}
			}
			\subfloat[ kodim09: 32.75dB]{
				\includegraphics[width=0.32\textwidth]{img/kodim/kodim09.png}
				\hspace{-3mm}
			}
		\end{minipage}
		\begin{minipage}{0.35\textwidth}
			\subfloat[ The 10 largest singular values in red channel of kodim08 and kodim03.]{
				\hspace{-1mm}
				\includegraphics[width=\textwidth]{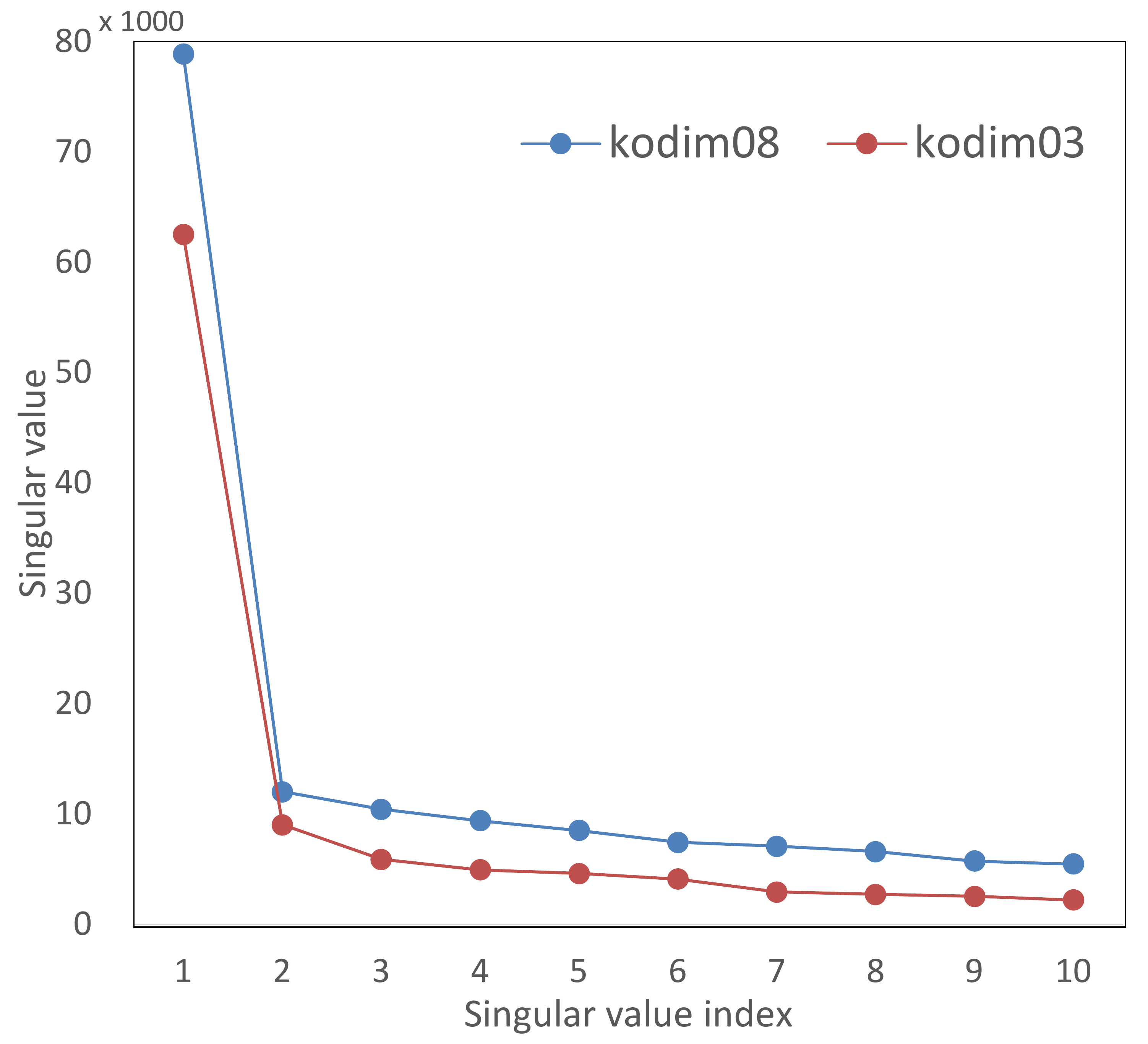}
				\label{top_10_sv}
			}
		\end{minipage}
		\caption{\centering The comparisons of the images with the 3 worst PSNR results (a $\sim$ c) and 3 best PSNR results (d $\sim$ f). And the comparison of the 10 largest singular values of kodim08 and kodim03.}
		\label{3-3}
	\end{figure*}

	\begin{table*}[thb] 
		\caption{\newline Grouping results of images in Kodak PhotoCD dataset.}
		\begin{tabular}{p{1.5cm}p{2cm}cccccccc|p{1cm}<{\centering}p{1cm}<{\centering}}
			\toprule
			\multirow{4}{*}{Group 1} & Image\# &13 &8 &5 &1 &24 &18 &14 &6 && \\
			& Avg PSNR &26.00 &27.77 &27.91 &28.20 &28.31 &28.40 &28.90 &29.49 && \\
			& Rank &1 &2 &3 &4 &5 &6 &7 &8 &\textbf{Avg} &\textbf{Std} \\
			&\textbf{Best} $\alpha$ &\textbf{2.40} &\textbf{2.10} &\textbf{1.95} &\textbf{1.75} &\textbf{1.85} &\textbf{1.65} &\textbf{1.65} &\textbf{1.70} &\textbf{1.88} &\textbf{0.26} \\
			\midrule
			\multirow{4}{*}{Group 2} & Image\# &21 &22 &11 &19 &2 &20 &17 &16 &&\\
			&Avg PSNR &29.66 &29.84 &29.97 &31.02 &31.35 &31.41 &31.51 &31.59 &&\\
			&Rank &9 &10 &11 &12 &13 &14 &15 &16 &\textbf{Avg} &\textbf{Std}\\
			&\textbf{Best $\alpha$} &\textbf{1.85} &\textbf{1.30} &\textbf{1.50} &\textbf{1.25} &\textbf{1.00} &\textbf{1.05} &\textbf{1.35} &\textbf{1.30} &\textbf{1.33} &\textbf{0.27}\\
			\midrule
			\multirow{4}{*}{Group 3} &Image\# &4 &15 &7 &10 &12 &9 &23 &3 &&\\
			&Avg PSNR &31.70 &31.81 &31.88 &32.31 &32.59 &32.75 &32.83 &33.50 &&\\
			&Rank &17 &18 &19 &20 &21 &22 &23 &24 &\textbf{Avg} &\textbf{Std}\\
			&\textbf{Best} $\alpha$ &\textbf{1.10} &\textbf{1.15} &\textbf{1.30} &\textbf{1.20} &\textbf{1.05} &\textbf{1.20} &\textbf{1.00} &\textbf{1.05} &\textbf{1.13} &\textbf{0.10}\\
			\bottomrule
		\end{tabular}
		\label{groups}
	\end{table*}
	\begin{table*}[thb] 
		\caption{\newline The table of ANVOA.}
		\begin{tabular}{p{2.5cm}p{2.7cm}<{\centering}p{3.5cm}<{\centering}p{2.5cm}<{\centering}p{1.3cm}<{\centering}c}
			\toprule
			{Source} &{Sum of Squares} &{Degrees of Freedom} &\makecell[l]{Mean Square} &$F$ &$p$ \\
			\midrule
			Inter-group &2.4252 &2 &1.2126 &24.38 &$\mathbf{3.35\times 10^{-6}}$ \\
			Within-group &1.0437 &21 &0.0497 && \\
			Total &3.4689 &23 &&& \\
			\bottomrule
		\end{tabular}
		\label{anova}
	\end{table*}
	
\subsection{ Analysis of parameter $\alpha$ }
	The setting of $\alpha$ is crucial to the effectiveness of the proposed model. Theoretically, $\alpha$ stems from the $L_1 - \alpha L_2$ norm \cite{L12}. In our model, $\alpha$ exists in the proximal operator $\mathbf{prox}_{ \lambda/\rho_k\Vert \cdot \Vert_{1-\alpha2} }$ (see section \textit{3.4} for detail) and will influence the shrinkage of different singular values, as shown in \figref{NNFN_alpha}. When $\alpha = 0$, the $L_1 - \alpha L_2$ norm reduces to the $L_1$ norm and hence the NNFN reduces to nuclear norm. In that case, $\mathbf{prox}_{ \lambda/\rho_k\Vert \cdot \Vert_{1-\alpha2} }$ is equivalent to the soft-thresholding operator \cite{L12}, which could result in the biased estimation. On the other side, when $\alpha \rightarrow +\infty$, NNFN only penalizes the several smallest singular values. And the largest and modest large singular values get zero penalty (i.e. be preserved). In such a case, the MC-NNFNM model might preserve too many noisy components while denoise little. 
	\par 
	Experimentally, we use the analysis of variance (ANOVA) to analyze the most suitable setting of $\alpha$ for the images with differen richness of edges, textures and colors. We test our model on Kodak PhotoCD dataset by changing $\alpha$ from 0.85 to 2.10 with an interval 0.05. Then we calculate the average PSNR for each image and sort them in ascending order. After that we divide the twenty-four images into three groups based on their average PSNR. As shown in Table \ref{groups}, the group 1, 2 and 3 contain the recovered images with the 8 lowest, 8 average and 8 highest PSNR results, respectively.
	\par
	The ANOVA $F$ test for ``best $\alpha$'' term is significant ($p < .0001$), as shown in Table \ref{anova}. This provides evidence that the best settings of $\alpha$ for three groups are not equal. Obviously, for those images achieving lower PSNR results (i.e. the images in group 1), a reletively lager $\alpha$ is preferred, and vice versa. \figref{3-3}, which includes 3 lowest and 3 highest PSNR results yield by our model, indicates that lower PSNR results are achieved by those images with complex edges, rich textures and indistinctive colors. The largest singular values of those images, as shown in \figref{top_10_sv}, contain more information and have larger magnitude. Hence a larger $\alpha$ is perferred to shrink them less or even preserve them.

\section{Conclusion}
	In this paper, a new low-rank minimization model was proposed to solve color image denoising problem. The proposed model has two major advantages. First, it can fully exploit the correlated information and noise difference among channels. Second, it satisfies adaptive shrinkage on singular values without assigning weights on them. With them, the proposed model is capable of achieving satisfactory results while keeping simplicity. An accurate and effective algorithm was designed to solve the proposed model based on ADMM framework. Moreover, rigorous convergence analysis and complexity analysis were presented to indicate the sound properties of the proposed model. Furthermore, ANOVA was resorted to discuss the influence of parameter $\alpha$ on denoising performance of the proposed model. Finally, the experimental results on synthetic and real noise datesets demonstrated the proposed model outperforms several state-of-the-art models.

	\appendix
	
	
	\bibliographystyle{elsarticle-num} 
	\bibliography{KBS23}

\begin{thebibliography}{10}
\expandafter\ifx\csname url\endcsname\relax
  \def\url#1{\texttt{#1}}\fi
\expandafter\ifx\csname urlprefix\endcsname\relax\def\urlprefix{URL }\fi
\expandafter\ifx\csname href\endcsname\relax
  \def\href#1#2{#2} \def\path#1{#1}\fi

\bibitem{iSegment1}
T.~Pappas, N.~Jayant, An adaptive clustering algorithm for image segmentation,
  in: International Conference on Acoustics, Speech, and Signal Processing,,
  1989, pp. 1667--1670 vol.3, https://doi.org/ICASSP.1989.266767.

\bibitem{iSegment2}
J.~Shi, J.~Malik, Normalized cuts and image segmentation, IEEE Transactions on
  Pattern Analysis and Machine Intelligence 22~(8) (2000) 888--905,
  https://doi.org/10.1109/34.868688.

\bibitem{RemoteSensingImaging1}
Y.~Chang, L.~Yan, T.~Wu, S.~Zhong, Remote sensing image stripe noise removal:
  From image decomposition perspective, IEEE Transactions on Geoscience and
  Remote Sensing 54~(12) (2016) 7018--7031, https://doi.org/TGRS.2016.2594080.

\bibitem{RemoteSensingImaging2}
G.~Bi, G.~Si, Y.~Zhao, B.~Qi, H.~Lv, Haze removal for a single remote sensing
  image using low-rank and sparse prior, IEEE Transactions on Geoscience and
  Remote Sensing 60 (2022) 1--13, https://doi.org/10.1109/TGRS.2021.3135975.

\bibitem{objRecognition}
P.-H. Hsiao, F.-J. Chang, Y.-Y. Lin, Learning discriminatively reconstructed
  source data for object recognition with few examples, IEEE Transactions on
  Image Processing 25~(8) (2016) 3518--3532,
  https://doi.org/10.1109/TIP.2016.2572602.

\bibitem{videoDenoising}
H.~Ji, C.~Liu, Z.~Shen, Y.~Xu, Robust video denoising using low rank matrix
  completion, in: 2010 IEEE Computer Society Conference on Computer Vision and
  Pattern Recognition, 2010, pp. 1791--1798,
  https://doi.org/10.1109/CVPR.2010.5539849.

\bibitem{BM3D}
K.~Dabov, A.~Foi, V.~Katkovnik, K.~Egiazarian, Image denoising by sparse 3-d
  transform-domain collaborative filtering, IEEE Transactions on Image
  Processing 16~(8) (2007) 2080--2095, https://doi.org/10.1109/TIP.2007.901238.

\bibitem{CBM3D}
K.~Dabov, A.~Foi, V.~Katkovnik, K.~Egiazarian, Color image denoising via sparse
  3d collaborative filtering with grouping constraint in luminance-chrominance
  space, in: 2007 IEEE International Conference on Image Processing, Vol.~1,
  2007, pp. I -- 313--I -- 316, https://doi.org/10.1109/ICIP.2007.4378954.

\bibitem{WNNM}
S.~Gu, L.~Zhang, W.~Zuo, X.~Feng, Weighted nuclear norm minimization with
  application to image denoising, in: 2014 IEEE Conference on Computer Vision
  and Pattern Recognition, 2014, pp. 2862--2869,
  https://doi.org/10.1109/CVPR.2014.366.

\bibitem{WSNM}
Y.~Xie, S.~Gu, Y.~Liu, W.~Zuo, W.~Zhang, L.~Zhang, Weighted schatten $p$-norm
  minimization for image denoising and background subtraction, IEEE
  Transactions on Image Processing 25~(10) (2016) 4842--4857,
  https://doi.org/10.1109/TIP.2016.2599290.

\bibitem{MCWNNM}
J.~Xu, L.~Zhang, D.~Zhang, X.~Feng, Multi-channel weighted nuclear norm
  minimization for real color image denoising, in: 2017 IEEE International
  Conference on Computer Vision (ICCV), 2017, pp. 1105--1113,
  https://doi.org/10.1109/ICCV.2017.125.

\bibitem{MCWSNM}
X.~Huang, B.~Du, W.~Liu, Multichannel color image denoising via weighted
  schatten p-norm minimization, in: C.~Bessiere (Ed.), Proceedings of the
  Twenty-Ninth International Joint Conference on Artificial Intelligence,
  {IJCAI-20}, International Joint Conferences on Artificial Intelligence
  Organization, 2020, pp. 637--644, main track.

\bibitem{DnCNN}
K.~Zhang, W.~Zuo, Y.~Chen, D.~Meng, L.~Zhang, Beyond a gaussian denoiser:
  Residual learning of deep cnn for image denoising, IEEE Transactions on Image
  Processing 26~(7) (2017) 3142--3155,
  https://doi.org/10.1109/TIP.2017.2662206.

\bibitem{FFDNet}
K.~Zhang, W.~Zuo, L.~Zhang, Ffdnet: Toward a fast and flexible solution for
  cnn-based image denoising, IEEE Transactions on Image Processing 27~(9)
  (2018) 4608--4622, https://doi.org/10.1109/TIP.2018.2839891.

\bibitem{wang2}
Z.~Wang, W.~Wang, J.~Wang, S.~Chen, Fast and efficient algorithm for matrix
  completion via closed-form 2/3-thresholding operator, Neurocomputing 330
  (2019) 212--222, https://doi.org/10.1016/j.neucom.2018.10.065.

\bibitem{wang3}
Z.~Wang, D.~Hu, X.~Luo, W.~Wang, J.~Wang, W.~Chen, Performance guarantees of
  transformed schatten-1 regularization for exact low-rank matrix recovery,
  International Journal of Machine Learning and Cybernetics 12~(12) (2021)
  3379--3395, https://doi.org/10.1007/s13042-021-01361-1.

\bibitem{wang4}
Z.~Wang, C.~Gao, X.~Luo, M.~Tang, J.~Wang, W.~Chen, Accelerated inexact matrix
  completion algorithm via closed-form q-thresholding $(q= 1/2, 2/3)$ operator,
  International Journal of Machine Learning and Cybernetics 11~(10) (2020)
  2327--2339, https://doi.org/10.1007/s13042-020-01121-7.

\bibitem{wang1}
Z.~Wang, Y.~Liu, X.~Luo, J.~Wang, C.~Gao, D.~Peng, W.~Chen, Large-scale affine
  matrix rank minimization with a novel nonconvex regularizer, IEEE
  Transactions on Neural Networks and Learning Systems 33~(9) (2022)
  4661--4675, https://doi.org/10.1109/TNNLS.2021.3059711.

\bibitem{tighest}
M.~Fazel, Matrix rank minimization with applications, Ph.D. thesis, PhD thesis,
  Stanford University (2002).

\bibitem{Candes}
E.~Cand\`{e}s, B.~Recht, Exact matrix completion via convex optimization,
  Commun. ACM 55~(6) (2012) 111–119, https://doi.org/10.1145/2184319.2184343.

\bibitem{SVT}
J.-F. Cai, E.~J. Cand\`{e}s, Z.~Shen, A singular value thresholding algorithm
  for matrix completion, SIAM Journal on Optimization 20~(4) (2010) 1956--1982.
\newblock \href {http://arxiv.org/abs/https://doi.org/10.1137/080738970}
  {\path{arXiv:https://doi.org/10.1137/080738970}}.

\bibitem{APGL}
A.~Beck, M.~Teboulle, A fast iterative shrinkage-thresholding algorithm for
  linear inverse problems, SIAM Journal on Imaging Sciences 2~(1) (2009)
  183--202.
\newblock \href {http://arxiv.org/abs/https://doi.org/10.1137/080716542}
  {\path{arXiv:https://doi.org/10.1137/080716542}}.

\bibitem{FPCA}
S.~Ma, D.~Goldfarb, L.~Chen, Fixed point and bregman iterative methods for
  matrix rank minimization, Mathematical Programming 128~(1) (2011) 321--353,
  https://doi.org/10.1007/s10107-009-0306-5.

\bibitem{SNM}
F.~Nie, H.~Huang, C.~Ding, Low-rank matrix recovery via efficient schatten
  p-norm minimization, in: Proceedings of the AAAI Conference on Artificial
  Intelligence, Vol.~26, 2012, pp. 655--661,
  https://doi.org/10.1609/aaai.v26i1.8210.

\bibitem{lq}
S.~Foucart, M.-J. Lai, Sparsest solutions of underdetermined linear systems via
  $\ell q$-minimization for $0<q \le 1$, Applied and Computational Harmonic
  Analysis 26~(3) (2009) 395--407, https://doi.org/10.1016/j.acha.2008.09.001.

\bibitem{TL12}
T.-H. Ma, Y.~Lou, T.-Z. Huang, Truncated $l\_{1-2}$ models for sparse recovery
  and rank minimization, SIAM Journal on Imaging Sciences 10~(3) (2017)
  1346--1380.
\newblock \href {http://arxiv.org/abs/https://doi.org/10.1137/16M1098929}
  {\path{arXiv:https://doi.org/10.1137/16M1098929}}.

\bibitem{capped_L1}
T.~Zhang, Analysis of multi-stage convex relaxation for sparse regularization,
  J. Mach. Learn. Res. 11 (2010) 1081–1107.

\bibitem{TNNR}
Y.~Hu, D.~Zhang, J.~Ye, X.~Li, X.~He, Fast and accurate matrix completion via
  truncated nuclear norm regularization, IEEE Transactions on Pattern Analysis
  and Machine Intelligence 35~(9) (2013) 2117--2130,
  https://doi.org/10.1109/TPAMI.2012.271.

\bibitem{MCP}
C.-H. Zhang, Nearly unbiased variable selection under minimax concave penalty,
  Annals of Statistics 38~(2) (2010) 894--942,
  https://doi.org/10.1214/09-AOS729.

\bibitem{nonconvex_better_theo}
R.~Mazumder, D.~Saldana, H.~Weng, Matrix completion with nonconvex
  regularization: Spectral operators and scalable algorithms, Statistics and
  Computing 30~(4) (2020) 1113--1138,
  https://doi.org/10.1007/s11222-020-09939-5.

\bibitem{nonconvex_better_emp}
Q.~Yao, J.~T. Kwok, T.~Wang, T.-Y. Liu, Large-scale low-rank matrix learning
  with nonconvex regularizers, IEEE transactions on pattern analysis and
  machine intelligence 41~(11) (2018) 2628--2643,
  https://doi.org/10.48550/arXiv.1708.00146.

\bibitem{MNLF}
J.~Dai, O.~C. Au, L.~Fang, C.~Pang, F.~Zou, J.~Li, Multichannel nonlocal means
  fusion for color image denoising, IEEE Transactions on Circuits and Systems
  for Video Technology 23~(11) (2013) 1873--1886,
  https://doi.org/10.1109/TCSVT.2013.2269020.

\bibitem{two_color_strategies}
F.~Luisier, T.~Blu, Sure-let multichannel image denoising: Interscale
  orthonormal wavelet thresholding, IEEE Transactions on Image Processing
  17~(4) (2008) 482--492, https://doi.org/10.1109/TIP.2008.919370.

\bibitem{sparRep4}
Z.~Kong, X.~Yang, Color image and multispectral image denoising using block
  diagonal representation, IEEE Transactions on Image Processing 28~(9) (2019)
  4247--4259, https://doi.org/10.1109/TIP.2019.2907478.

\bibitem{concatenate}
M.~Lebrun, M.~Colom, J.-M. Morel, Multiscale image blind denoising, IEEE
  Transactions on Image Processing 24~(10) (2015) 3149--3161,
  https://doi.org/10.1109/TIP.2015.2439041.

\bibitem{spatial_spectral}
P.~Zhong, R.~Wang, Multiple-spectral-band crfs for denoising junk bands of
  hyperspectral imagery, IEEE Transactions on Geoscience and Remote Sensing
  51~(4) (2013) 2260--2275, https://doi.org/10.1109/TGRS.2012.2209656.

\bibitem{NNFN}
Y.~Wang, Q.~Yao, J.~Kwok, A scalable, adaptive and sound nonconvex regularizer
  for low-rank matrix learning, WWW '21, Association for Computing Machinery,
  New York, NY, USA, 2021, p. 1798–1808,
  https://doi.org/10.1145/3442381.3450142.
\newblock \href {https://doi.org/10.1145/3442381.3450142}
  {\path{doi:10.1145/3442381.3450142}}.

\bibitem{ADMM}
S.~Boyd, N.~Parikh, E.~Chu, B.~Peleato, J.~Eckstein, Distributed optimization
  and statistical learning via the alternating direction method of multipliers,
  Foundations and Trends® in Machine Learning 3~(1) (2011) 1--122,
  http://dx.doi.org/10.1561/2200000016.
\newblock \href {https://doi.org/10.1561/2200000016}
  {\path{doi:10.1561/2200000016}}.

\bibitem{convex_conv1}
A.~Ruszczy{\'n}ski, On convergence of an augmented lagrangian decomposition
  method for sparse convex optimization, Mathematics of Operations Research
  20~(3) (1995) 634--656, https://doi.org/10.1287/moor.20.3.634.

\bibitem{convex_conv3}
D.~Davis, Convergence rate analysis of primal-dual splitting schemes, SIAM
  Journal on Optimization 25~(3) (2015) 1912--1943,
  https://doi.org/10.48550/arXiv.1408.4419.

\bibitem{noncon_conv2}
Y.~Wang, W.~Yin, J.~Zeng, Global convergence of admm in nonconvex nonsmooth
  optimization, Journal of Scientific Computing 78~(1) (2019) 29--63,
  https://doi.org/10.48550/arXiv.1511.06324.

\bibitem{ADMM_app_nonc1}
Y.~Xu, W.~Yin, Z.~Wen, Y.~Zhang, An alternating direction algorithm for matrix
  completion with nonnegative factors, Frontiers of Mathematics in China 7~(2)
  (2012) 365--384, https://doi.org/10.1007/s11464-012-0194-5.

\bibitem{ADMM_app_nonc2}
C.~Schenker, J.~E. Cohen, E.~Acar, An optimization framework for regularized
  linearly coupled matrix-tensor factorization, in: 2020 28th European Signal
  Processing Conference (EUSIPCO), 2021, pp. 985--989,
  https://doi.org/10.23919/Eusipco47968.2020.9287459.

\bibitem{NSP}
S.~Wang, L.~Zhang, Y.~Liang, Nonlocal spectral prior model for low-level
  vision, in: K.~M. Lee, Y.~Matsushita, J.~M. Rehg, Z.~Hu (Eds.), Computer
  Vision -- ACCV 2012, Springer Berlin Heidelberg, Berlin, Heidelberg, 2013,
  pp. 231--244, https://doi.org/10.1007/978-3-642-37431-9\_18.

\bibitem{GST}
W.~Zuo, D.~Meng, L.~Zhang, X.~Feng, D.~Zhang, A generalized iterated shrinkage
  algorithm for non-convex sparse coding, in: 2013 IEEE International
  Conference on Computer Vision, 2013, pp. 217--224,
  https://doi.org/10.1109/ICCV.2013.34.

\bibitem{Assume_AWGN}
B.~Leung, G.~Jeon, E.~Dubois, Least-squares luma–chroma demultiplexing
  algorithm for bayer demosaicking, IEEE Transactions on Image Processing
  20~(7) (2011) 1885--1894, https://doi.org/10.1109/TIP.2011.2107524.

\bibitem{L12}
Y.~Lou, M.~Yan, Fast l1--l2 minimization via a proximal operator, Journal of
  Scientific Computing 74~(2) (2018) 767--785,
  https://doi.org/10.48550/arXiv.1609.09530.

\bibitem{GID}
J.~Xu, L.~Zhang, D.~Zhang, External prior guided internal prior learning for
  real-world noisy image denoising, IEEE Transactions on Image Processing
  27~(6) (2018) 2996--3010, https://doi.org/10.1109/TIP.2018.2811546.

\bibitem{NI}
Neatlab, Neat image, \url{https://ni.neatvideo.com/home} (2022).

\bibitem{noisEstimate}
G.~Chen, F.~Zhu, P.-A. Heng, An efficient statistical method for image noise
  level estimation, 2015 IEEE International Conference on Computer Vision
  (ICCV) (2015) 477--485Https://doi.org/10.1109/ICCV.2015.62.

\bibitem{CC15}
S.~Nam, Y.~Hwang, Y.~Matsushita, S.~J. Kim, A holistic approach to
  cross-channel image noise modeling and its application to image denoising,
  in: 2016 IEEE Conference on Computer Vision and Pattern Recognition (CVPR),
  2016, pp. 1683--1691, https://doi.org/10.1109/CVPR.2016.186.

\end{thebibliography}
	
	
	%
	%
	%
\end{document}